\documentclass[runningheads]{llncs}

\usepackage{graphicx}
\usepackage{tikz}
\usepackage{comment}
\usepackage{amsmath,amssymb} % define this before the line numbering.
\usepackage{color}
\usepackage{pifont}
\usepackage{booktabs}
\usepackage{commath}
\usepackage[caption=false]{subfig}

\usepackage[accsupp]{axessibility}  % Improves PDF readability for those with disabilities.

\usepackage[pagebackref,breaklinks,colorlinks]{hyperref}

\def\Approach{Counting-DETR}

\def\block1{{$\sf block1$}}
\def\block2{{$\sf block2$}}
\def\block3{{$\sf block3$}}
\def\block4{{$\sf block4$}}

\def\R{{\mathbb R}}

\definecolor{mydarkblue}{rgb}{0,0.08,1}
\definecolor{mydarkgreen}{rgb}{0.02,0.6,0.02}
\definecolor{myred}{rgb}{1.0,0.0,0.0}

\begin{document}
\def\mA{\mathcal{A}}
\def\mB{\mathcal{B}}
\def\mC{\mathcal{C}}
\def\mD{\mathcal{D}}
\def\mE{\mathcal{E}}
\def\mF{\mathcal{F}}
\def\mG{\mathcal{G}}
\def\mH{\mathcal{H}}
\def\mI{\mathcal{I}}
\def\mJ{\mathcal{J}}
\def\mK{\mathcal{K}}
\def\mL{\mathcal{L}}
\def\mM{\mathcal{M}}
\def\mN{\mathcal{N}}
\def\mO{\mathcal{O}}
\def\mP{\mathcal{P}}
\def\mQ{\mathcal{Q}}
\def\mR{\mathcal{R}}
\def\mS{\mathcal{S}}
\def\mT{\mathcal{T}}
\def\mU{\mathcal{U}}
\def\mV{\mathcal{V}}
\def\mW{\mathcal{W}}
\def\mX{\mathcal{X}}
\def\mY{\mathcal{Y}}
\def\mZ{\mathcal{Z}} 

\def\bbN{\mathbb{N}} 
\def\bbR{\mathbb{R}} 
\def\bbP{\mathbb{P}} 
\def\bbQ{\mathbb{Q}} 
\def\bbE{\mathbb{E}}

\def\1n{\mathbf{1}_n}
\def\0{\mathbf{0}}
\def\1{\mathbf{1}}

\def\A{{\bf A}}
\def\B{{\bf B}}
\def\C{{\bf C}}
\def\D{{\bf D}}
\def\E{{\bf E}}
\def\F{{\bf F}}
\def\G{{\bf G}}
\def\H{{\bf H}}
\def\I{{\bf I}}
\def\J{{\bf J}}
\def\K{{\bf K}}
\def\L{{\bf L}}
\def\M{{\bf M}}
\def\N{{\bf N}}
\def\O{{\bf O}}
\def\P{{\bf P}}
\def\Q{{\bf Q}}
\def\R{{\bf R}}
\def\S{{\bf S}}
\def\T{{\bf T}}
\def\U{{\bf U}}
\def\V{{\bf V}}
\def\W{{\bf W}}
\def\X{{\bf X}}
\def\Y{{\bf Y}}
\def\Z{{\bf Z}}

\def\a{{\bf a}}
\def\b{{\bf b}}
\def\c{{\bf c}}
\def\d{{\bf d}}
\def\e{{\bf e}}
\def\f{{\bf f}}
\def\g{{\bf g}}
\def\h{{\bf h}}
\def\i{{\bf i}}
\def\j{{\bf j}}
\def\k{{\bf k}}
\def\l{{\bf l}}
\def\m{{\bf m}}
\def\n{{\bf n}}
\def\o{{\bf o}}
\def\p{{\bf p}}
\def\q{{\bf q}}
\def\r{{\bf r}}
\def\s{{\bf s}}
\def\t{{\bf t}}
\def\u{{\bf u}}
\def\v{{\bf v}}
\def\w{{\bf w}}
\def\x{{\bf x}}
\def\y{{\bf y}}
\def\z{{\bf z}}

\def\balpha{\mbox{\boldmath{$\alpha$}}}
\def\bbeta{\mbox{\boldmath{$\beta$}}}
\def\bdelta{\mbox{\boldmath{$\delta$}}}
\def\bgamma{\mbox{\boldmath{$\gamma$}}}
\def\blambda{\mbox{\boldmath{$\lambda$}}}
\def\bsigma{\mbox{\boldmath{$\sigma$}}}
\def\btheta{\mbox{\boldmath{$\theta$}}}
\def\bomega{\mbox{\boldmath{$\omega$}}}
\def\bxi{\mbox{\boldmath{$\xi$}}}
\def\bnu{\mbox{\boldmath{$\nu$}}}                                  
\def\bphi{\mbox{\boldmath{$\phi$}}}
\def\bmu{\mbox{\boldmath{$\mu$}}}

\def\bDelta{\mbox{\boldmath{$\Delta$}}}
\def\bOmega{\mbox{\boldmath{$\Omega$}}}
\def\bPhi{\mbox{\boldmath{$\Phi$}}}
\def\bLambda{\mbox{\boldmath{$\Lambda$}}}
\def\bSigma{\mbox{\boldmath{$\Sigma$}}}
\def\bGamma{\mbox{\boldmath{$\Gamma$}}}
                                  
\newcommand{\myprob}[1]{\mathop{\mathbb{P}}_{#1}}

\newcommand{\myexp}[1]{\mathop{\mathbb{E}}_{#1}}

\newcommand{\mydelta}[1]{1_{#1}}

\newcommand{\myminimum}[1]{\mathop{\textrm{minimum}}_{#1}}
\newcommand{\mymaximum}[1]{\mathop{\textrm{maximum}}_{#1}}    
\newcommand{\mymin}[1]{\mathop{\textrm{minimize}}_{#1}}
\newcommand{\mymax}[1]{\mathop{\textrm{maximize}}_{#1}}
\newcommand{\mymins}[1]{\mathop{\textrm{min.}}_{#1}}
\newcommand{\mymaxs}[1]{\mathop{\textrm{max.}}_{#1}}  
\newcommand{\myargmin}[1]{\mathop{\textrm{argmin}}_{#1}} 
\newcommand{\myargmax}[1]{\mathop{\textrm{argmax}}_{#1}} 
\newcommand{\myst}{\textrm{s.t. }}

\newcommand{\denselist}{\itemsep -1pt}
\newcommand{\sparselist}{\itemsep 1pt}

\definecolor{pink}{rgb}{0.9,0.5,0.5}
\definecolor{purple}{rgb}{0.5, 0.4, 0.8}   
\definecolor{gray}{rgb}{0.3, 0.3, 0.3}
\definecolor{mygreen}{rgb}{0.2, 0.6, 0.2}

\newcommand{\cyan}[1]{\textcolor{cyan}{#1}}
\newcommand{\red}[1]{\textcolor{red}{#1}}  
\newcommand{\blue}[1]{\textcolor{blue}{#1}}
\newcommand{\magenta}[1]{\textcolor{magenta}{#1}}
\newcommand{\pink}[1]{\textcolor{pink}{#1}}
\newcommand{\green}[1]{\textcolor{green}{#1}} 
\newcommand{\gray}[1]{\textcolor{gray}{#1}}    
\newcommand{\mygreen}[1]{\textcolor{mygreen}{#1}}    
\newcommand{\purple}[1]{\textcolor{purple}{#1}}       

\definecolor{greena}{rgb}{0.4, 0.5, 0.1}
\newcommand{\greena}[1]{\textcolor{greena}{#1}}

\definecolor{bluea}{rgb}{0, 0.4, 0.6}
\newcommand{\bluea}[1]{\textcolor{bluea}{#1}}
\definecolor{reda}{rgb}{0.6, 0.2, 0.1}
\newcommand{\reda}[1]{\textcolor{reda}{#1}}

\def\changemargin#1#2{\list{}{\rightmargin#2\leftmargin#1}\item[]}
\let\endchangemargin=\endlist
                                               
\newcommand{\cm}[1]{}

\newcommand{\mhoai}[1]{{\color{magenta}\textbf{[MH: #1]}}}

\newcommand{\mtodo}[1]{{\color{red}$\blacksquare$\textbf{[TODO: #1]}}}
\newcommand{\myheading}[1]{\vspace{1ex}\noindent \textbf{#1}}
\newcommand{\htimesw}[2]{\mbox{$#1$$\times$$#2$}}

% The following are useful for creating homework or exams

\newif\ifshowsolution
%\showsolutionfalse
\showsolutiontrue

\ifshowsolution  
\newcommand{\Comment}[1]{\paragraph{\bf $\bigstar $ COMMENT:} {\sf #1} \bigskip}
\newcommand{\Solution}[2]{\paragraph{\bf $\bigstar $ SOLUTION:} {\sf #2} }
\newcommand{\Mistake}[2]{\paragraph{\bf $\blacksquare$ COMMON MISTAKE #1:} {\sf #2} \bigskip}
\else
\newcommand{\Solution}[2]{\vspace{#1}}
\fi

\newcommand{\truefalse}{
\begin{enumerate}
	\item True
	\item False
\end{enumerate}
}

\newcommand{\yesno}{
\begin{enumerate}
	\item Yes
	\item No
\end{enumerate}
}

\newcommand{\Sref}[1]{Sec.~\ref{#1}}
\newcommand{\Eref}[1]{Eq.~(\ref{#1})}
\newcommand{\Fref}[1]{Fig.~\ref{#1}}
\newcommand{\Tref}[1]{Table~\ref{#1}}

% \linenumbers
\pagestyle{headings}
\mainmatter
\def\ECCVSubNumber{5138}  % Insert your submission number here

\title{Few-shot Object Counting and Detection} % Replace with your title

\begin{comment}
\titlerunning{ECCV-22 submission ID \ECCVSubNumber} 
\authorrunning{ECCV-22 submission ID \ECCVSubNumber} 
\author{Anonymous ECCV submission}
\institute{Paper ID \ECCVSubNumber}
\end{comment}
%******************

% \begin{comment}
\titlerunning{Few-shot Object Counting and Detection}
\author{Thanh Nguyen\inst{1}\thanks{Equal contribution} \and
Chau Pham\inst{1}$^\star$ \and
Khoi Nguyen\inst{1} \and
Minh Hoai\inst{1, 2}}
\authorrunning{Nguyen \textsl{et al.}}
\institute{VinAI Research, Hanoi, Vietnam, \and
Stony Brook University, Stony Brook, NY, USA.} 
% \email{lncs@springer.com}
% \end{comment}
%******************
\maketitle

\newcommand{\xmark}{\ding{55}}

\begin{abstract}

We tackle a new task of few-shot object counting and detection. Given a few exemplar bounding boxes of a target object class, we seek to count and detect all objects of the target class. This task shares the same supervision as the few-shot object counting but additionally outputs the object bounding boxes along with the total object count. 
To address this challenging problem, we introduce a novel two-stage training strategy and a novel uncertainty-aware few-shot object detector: \Approach. The former is aimed at generating pseudo ground-truth bounding boxes to train the latter. The latter leverages the pseudo ground-truth provided by the former but takes the necessary steps to account for the imperfection of pseudo ground-truth. 
To validate the performance of our method on the new task, we introduce two new datasets named FSCD-147 and FSCD-LVIS. Both datasets contain images with complex scenes, multiple object classes per image, and a huge variation in object shapes, sizes, and appearance. Our proposed approach outperforms very strong baselines adapted from few-shot object counting and few-shot object detection with a large margin in both counting and detection metrics. The code and models are available at \url{https://github.com/VinAIResearch/Counting-DETR}.

% This problem is useful in the application where the bounding box is easier to verify the counting results than density map; or in the data annotation where the need of automatically annotate the bounding boxes of the new object class given a few exemplars. 
% Most existing visual object counting methods only output a total count number and a real-value density map. However, in many real-world applications, we need both the object counting and the detection such as in data. Furthermore, knowing the locations of the counted objects will also facilitate easy verification of the counting results.
% adapt Anchor DETR with two stage training with uncertainty aware loss. Our two stage training ease the learning process by generate pseudo label from very few exemplar boxes and annotated points then use generated label to train second stage. Our novel uncertainty aware loss tackle the imperfection of pseudo label from first stage. We also provide two new datasets to evaluate counting and detection concurrently. Our FSCD-147 dataset has 2476 fully annotated image with box and dot annotation in the test set and val set, which serve as a benchmark for this task. Our FSCD-LVIS contain more than 6000 images with even more complexity in scene, size and shape. Our source code and dataset will be publicly available.
% \dots

\keywords{Few-shot Object Counting, Few-shot Object Detection}
\end{abstract}

\section{Introduction}

\begin{figure}[t]
  \subfloat[Training]{
	\begin{minipage}[c]{
	   0.47\textwidth}
	   \centering
	   \includegraphics[width=1\textwidth]{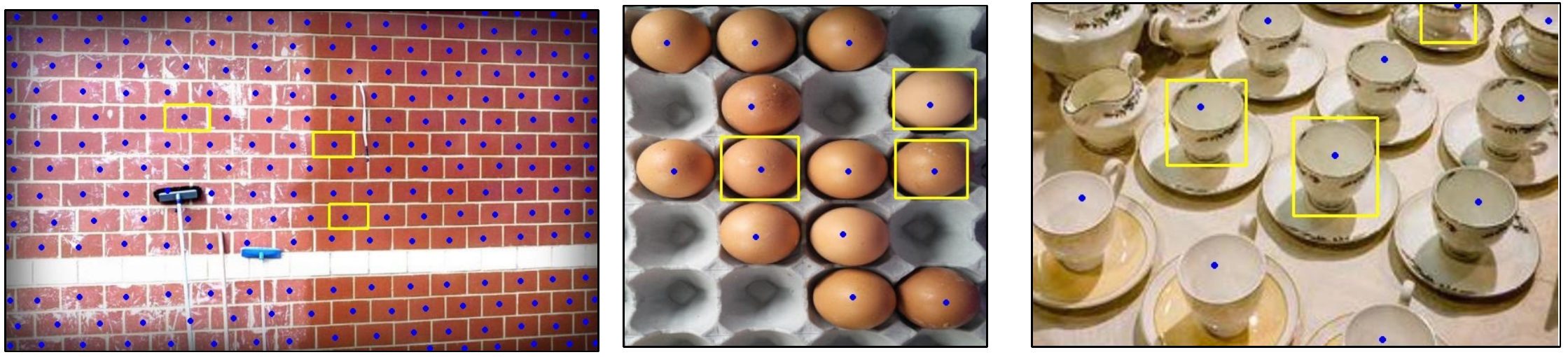}
	\end{minipage}} \hfill 
  \subfloat[Testing]{
	\begin{minipage}[c]{
	   0.47\textwidth}
	   \centering
	   \includegraphics[width=0.92\textwidth]{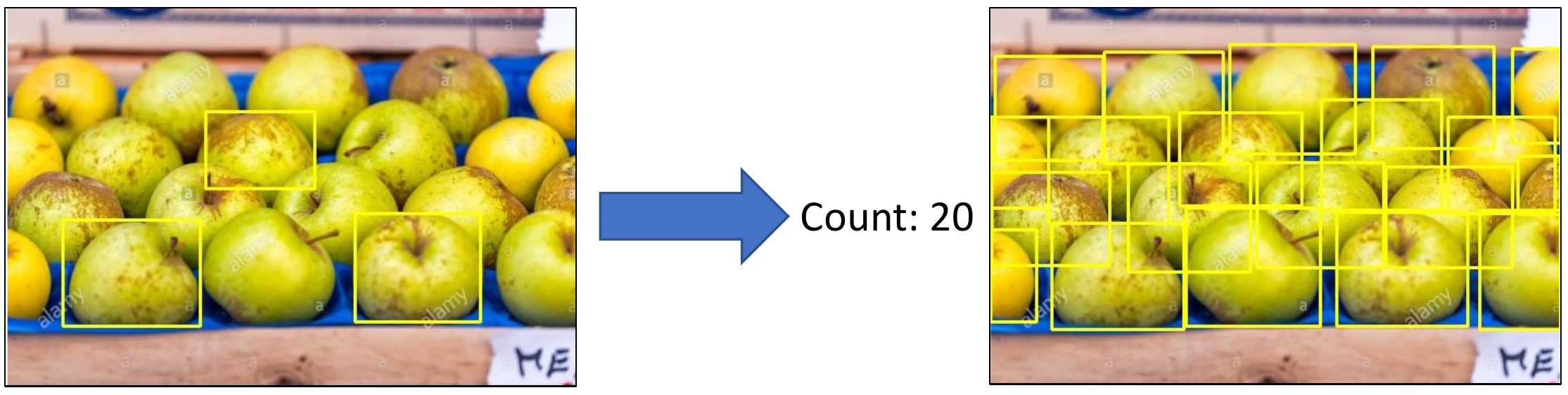}
	\end{minipage}}
\vspace{-5pt}
\caption{We address the task of few-shot counting and detection in a novel setting: (a) in training, each training image contains dot annotations for all objects and a few exemplar boxes. (b) In testing, given an image with a few exemplar boxes defining a target class, our goal is to count and detect all objects of that target class in the image.}
\label{fig:problem_statement}
\vspace{-10pt}
\end{figure}

This paper addresses a new task of Few-Shot object Counting and Detection (FSCD) in crowded scenes. Given an image containing many objects of multiple classes, we seek to count and detect all objects of a target class of interest specified by a few exemplar bounding boxes in the image. To facilitate few-shot learning, in training, we are only given the supervision of few-shot object counting, i.e., dot annotations for the approximate centers of all objects  and a few exemplar bounding boxes for object instances from the target class. It is worth noting that the test classes may or may not be present in training classes. The problem setting is depicted in Fig.~\ref{fig:problem_statement}.

FSCD is different from Few-Shot Object Counting (FSC) and Few-Shot Object Detection (FSOD). Compared to FSC, FSCD has several advantages: (1) obtaining object bounding boxes ``for free'', which is suitable for quickly annotating bounding boxes for a new object class with a few exemplar bounding boxes; (2) making the result of FSC more interpretable since bounding boxes are easier to verify than the density map. Compared to FSOD which requires bounding box annotation for all objects in the training phase of the base classes, FSCD uses significantly less supervision, i.e., only a few exemplar bounding boxes and dot annotations for all objects. This is helpful in crowded scenes where annotating accurate bounding boxes for all objects is ambiguously harder and significantly more expensive than the approximate dot annotation. 

Consequently, FSCD is more challenging than both FSC and FSOD. FSCD needs to detect and count all the objects as for FSOD, but it is only trained with the supervision of FSC. This invalidates of most of available approaches used in these problems without significant changes in network architecture or loss function. Specifically, it is not trivial to extend the density map produced by FSC approaches to predict the object bounding boxes; and it is hard to train a few-shot object detector with few exemplar bounding boxes of the base classes.

%In particular, to extend a current dataset (the FSC-147 dataset \cite{m_Ranjan-etal-CVPR21}) for FSOD in crowded scene to FSCD, we additionally annotate bounding box annotations for the validation set of 1190 images, it took us around 420 hours compared to annotate the dot annotation of that set in less than 60 hours. 

% \begin{figure}
% \centering
% \begin{subfigure}{.5\textwidth}
%   \centering
%   \includegraphics[width=\linewidth, left]{images/problem_statement_training.pdf}
%   \caption{Training}
% %   \label{fig:baseline}
% \end{subfigure}%
% \begin{subfigure}{.5\textwidth}
%   \centering
%   \includegraphics[width=.95\linewidth, right]{images/problem_statement_test.pdf}
%   \caption{Testing}
% %   \label{fig:common_errors_1}
% \end{subfigure}
% \caption{The FSCD problem setting. (a) In training we are given a images with dot annotations for all objects and a few exemplar boxes. (b) In testing, we given an image with a few exemplar boxes defining the target class, our goal is to count and detect all objects of that target class in the image.}
% \label{fig:problem_statement}
% \end{figure}

\begin{figure}[ht]
  \subfloat[]{
	\begin{minipage}[c]{
	   .3\textwidth}
	   \centering
	   \label{fig:baseline}
	   \includegraphics[width=1\textwidth]{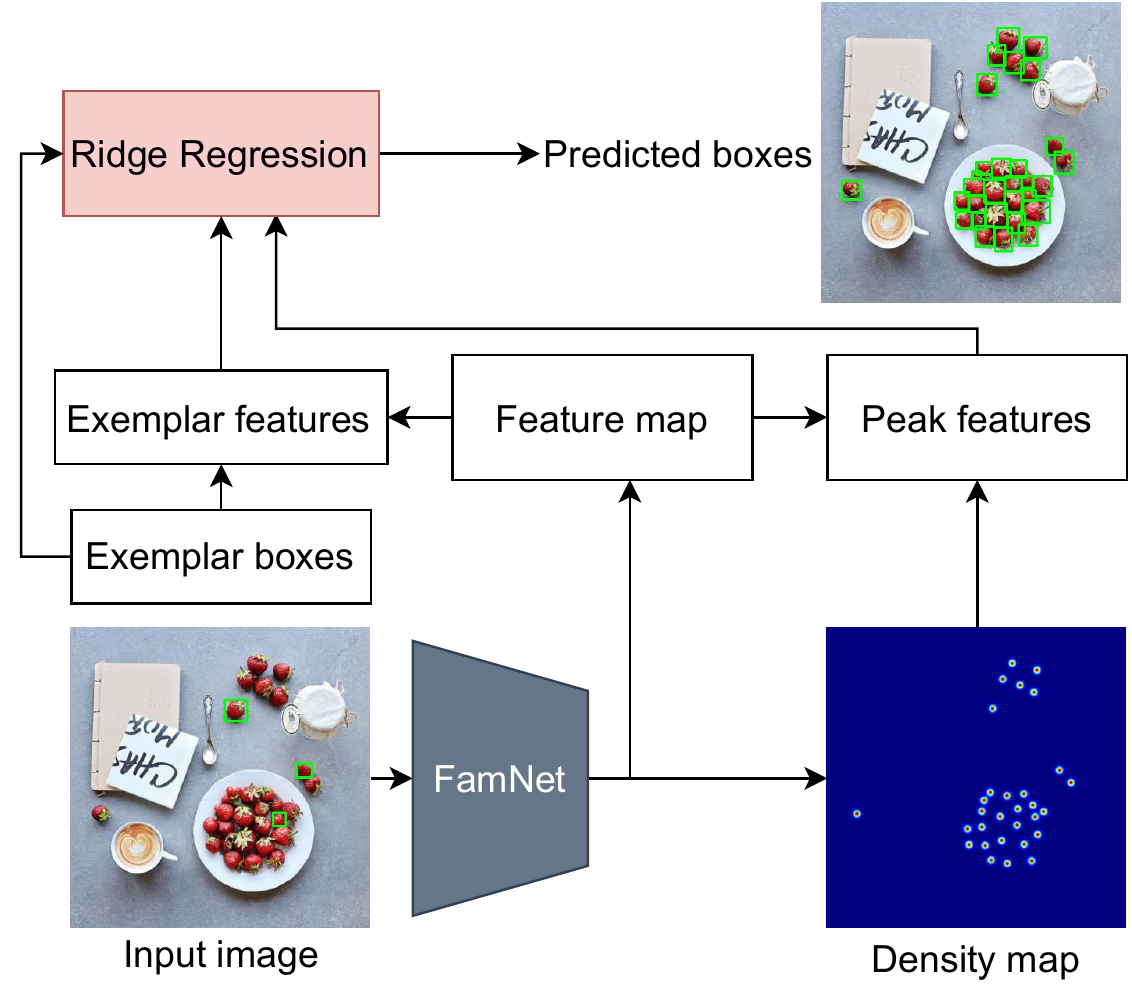}
	\end{minipage}}
  \subfloat[]{
	\begin{minipage}[c]{
	   0.39\textwidth}
	   \centering
	   \label{fig:common_errors_1}
	   \includegraphics[width=0.95\textwidth]{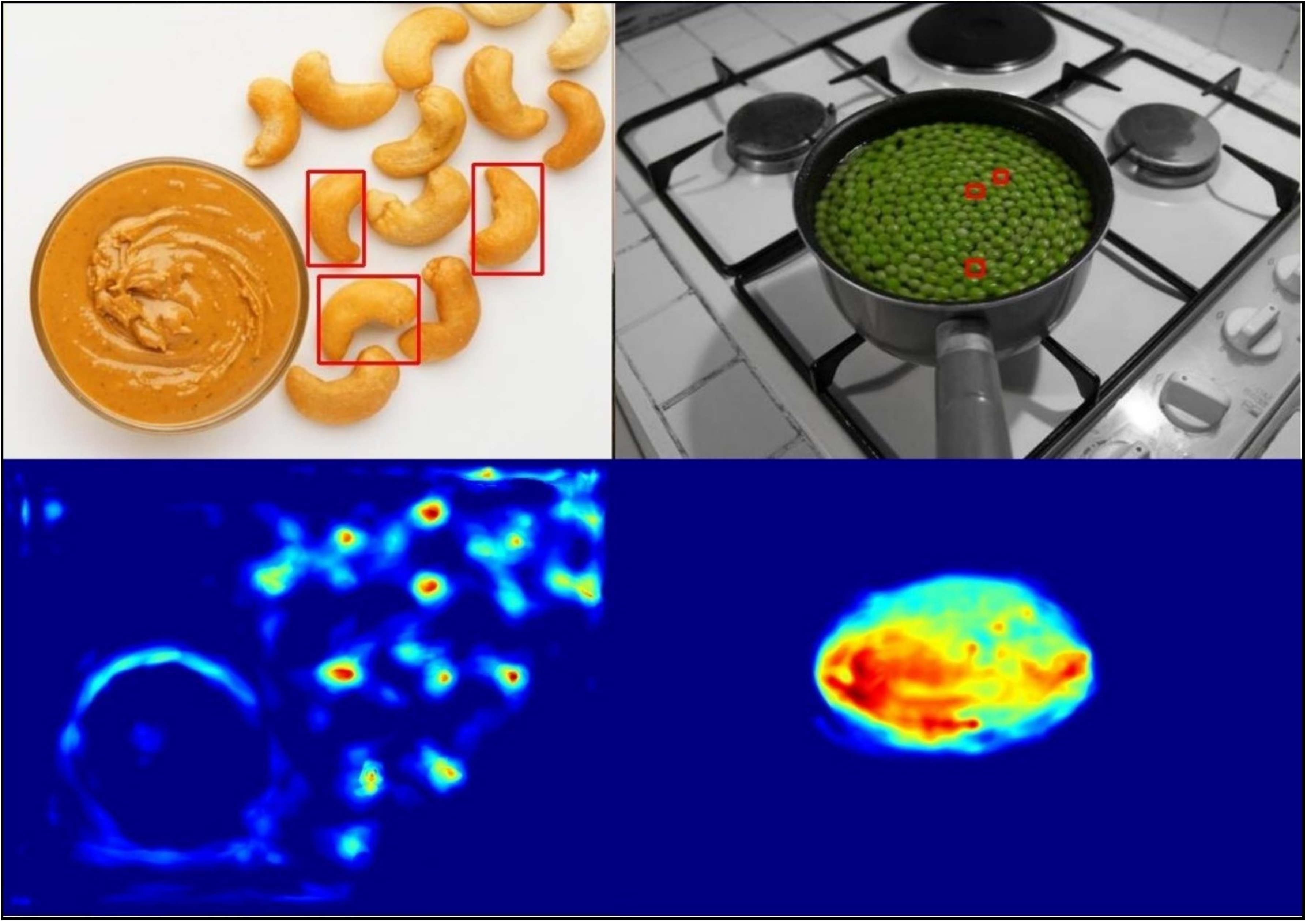}
	\end{minipage}}
   \subfloat[]{
	\begin{minipage}[c]{
	   0.3\textwidth}
	   \centering
	   \label{fig:common_errors_2}
	   \includegraphics[width=0.95\textwidth]{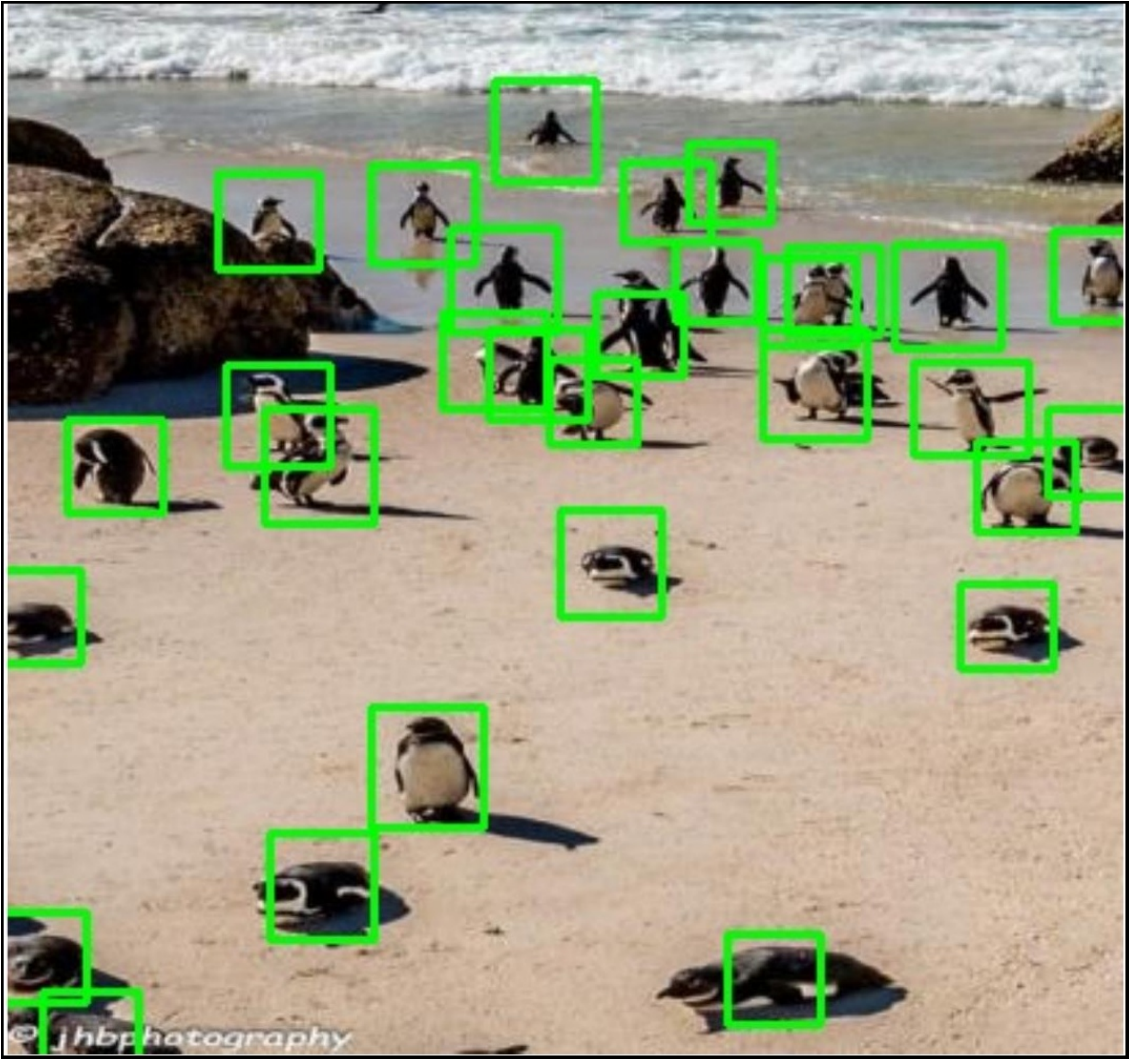}
	\end{minipage}}
\vspace{-5pt}
\caption{Limitations of a naive approach for FSCD by extending FamNet \cite{m_Ranjan-etal-CVPR21} with a regression function for object detection. (a) Processing pipeline of this approach: a regressor takes as input exemplar boxes with their features, and features at peak density locations to predict bounding boxes for the peak locations. (b) Limitation 1: poor quality of the density map predicted by FamNet when the exemplars share similar appearance with background or densely packed region. The first row presents the input images with a few exemplars each, the second row presents the corresponding density map predicted by FamNet. (c) Limitation 2: Non-discriminative peak features cannot represent objects with significant differences in shape and size. The green boxes are predicted from the features extracted at the annotated dots. }
\label{fig:famnet}
\vspace{-10pt}
\end{figure}

A naive approach for FSCD is to extend FamNet \cite{m_Ranjan-etal-CVPR21}, a density-map-based approach for FSC, whose counting number is obtained by summing over the predicted density map. To extend FamNet to detect objects, one can use a regression function on top of the features extracted from the peak locations (whose density values are highest in their respective local neighborhoods), the features extracted from the exemplars, and the exemplar boxes themselves. The process of this naive approach is illustrated in \Fref{fig:baseline}.
However, this approach has two limitations due to: 1) the imperfection of the predicted density map, and 2) the non-discriminative peak features. In the former, the density value is high in the environment locations whose color is similar to those of the exemplars, or the density map is peak-indistinguishable when the objects are packed in a dense region as depicted in \Fref{fig:common_errors_1}. In the latter, the extracted features are trained with counting objective (not object detection) so that they cannot represent for different shapes, sizes, and orientations, as illustrated in \Fref{fig:common_errors_2}.

% A naive approach for FSCD is to extend FamNet \cite{m_Ranjan-etal-CVPR21}, a density-map-based approach for FSC, whose counting number is obtained by summing over the predicted density map. To extend FamNet to detect objects, one can use Ridge Regression on top of the features extracted from the peak locations (whose density values are higher than the local neighborhood), the features extracted from the exemplars, and the exemplar boxes themselves. The whole process is illustrated in \Fref{fig:baseline}.
% Ridge regression has several advantages including a closed-form solution for fast training and regularization capability for not overfitting to
% a few training examples. Also, ridge regression is highly suitable for the few-shot setting, as proposed by \cite{bertinetto2018meta} for few-shot image classification. However, this approach has many limitations due to the imperfection of the predicted density map and the non-discriminative peak features. In the former, the density value is high in the environment locations whose color is similar to those of the exemplars, or the density map is peak-indistinguishable when the objects are packed in a dense region as depicted in \Fref{fig:common_errors_1}. In the latter, the extracted features are trained with counting objective (not object detection) so that they cannot represent for different shapes, sizes, and orientations, as illustrated in \Fref{fig:common_errors_2}.

\begin{figure}[t!]
\centering
\includegraphics[scale=0.47]{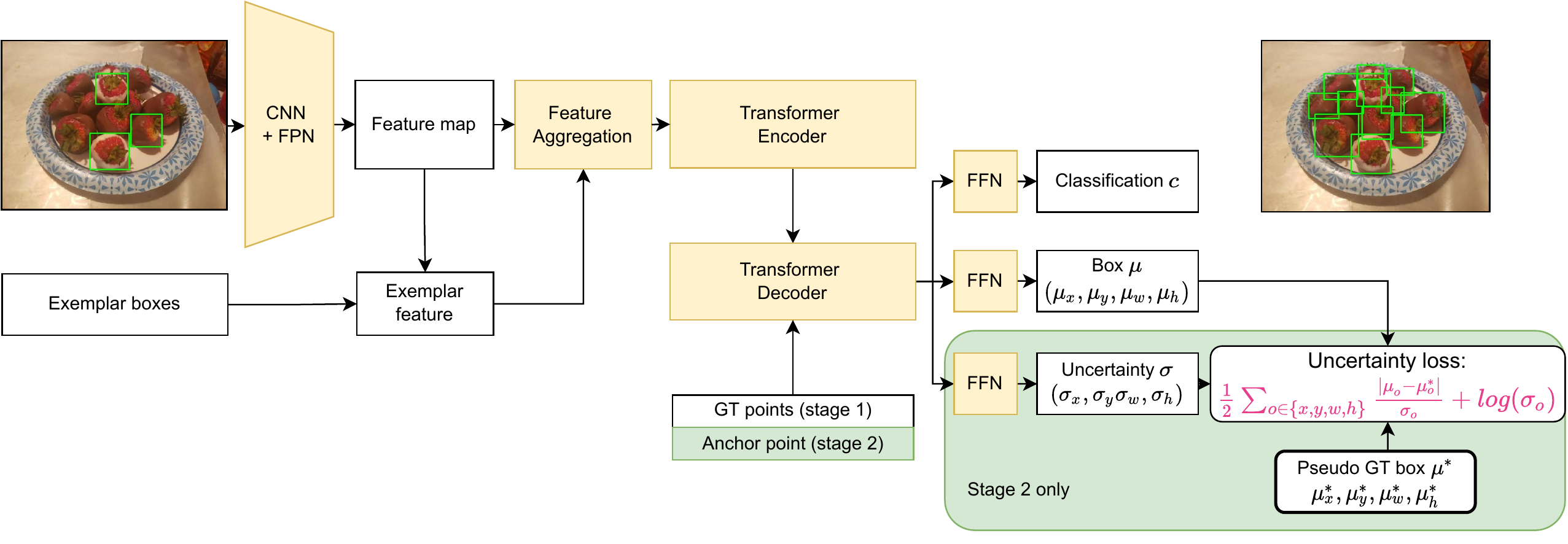}
%\vspace{-5pt}
\vskip -0.1in
\caption{The overview of our two-step training approach: (1) \Approach~is first trained on a few pairs of dot and bounding boxes and then used to predict pseudo GT boxes for the annotated dots; (2) \Approach~is trained to predict the object bounding boxes, with the prediction target being the pseudo GT boxes from the first stage. Specifically, the input image is first forwarded through a CNN+FPN backbone to extract its feature map. The exemplar features are extracted from their boxes to integrate with the feature map producing the exemplars-integrated feature map. This feature map is then taken as input to the encoder-decoder transformer along with either the annotated dots in the first stage or the anchor points in the second stage for foreground/background classification and bounding box regression.  In the second stage, the estimated uncertainty is used to regularize the training with a new uncertainty loss to account for the imperfection of the pseudo GT bounding boxes.}
\label{fig:main_diagram}
%\vspace{-15pt}
\end{figure}

% Put the description of our Approach to address the above limitations
To address the aforementioned limitations, we propose a new point-based approach, named \Approach, treating objects as points. In particular, counting and detecting objects is equivalent to counting and detecting points, and the object bounding box is predicted directly from point features. 
% Our approach overcomes the limitation of density-map-based and detection-based approaches. 
\Approach~is based on an object detector, Anchor DETR \cite{wang2021anchor}, with  improvements to better address FSCD. 
\textbf{First}, inspired by \cite{chen2021points} we adopt a two-stage training strategy: (1) \Approach~is trained to generate pseudo ground-truth (GT) bounding boxes given the annotated points of training images; (2) \Approach~is further fine-tuned on the generated pseudo GT bounding boxes to detect objects on test images. \textbf{Second,} since the generated pseudo GT bounding boxes are imperfect, we propose to estimate the uncertainty for bounding box prediction in the second stage. The estimated uncertainty regularizes learning such that lower box regression loss is incurred on the predictions with high uncertainty.
The overview of \Approach~is illustrated in Fig.~\ref{fig:main_diagram}.

% \Approach~outperforms a strong baseline in these datasets. 
% describe more the result in here. I will add result later.

In short, the contributions of our paper are: (1) we introduce a new problem of few-shot object counting and detection (FSCD); (2) we introduce two new datasets, FSCD-147 and FSCD-LVIS; (3) we propose a two-stage training strategy to first generate pseudo GT bounding boxes from the dot annotations, then use these boxes as supervision for training our proposed few-shot object detector; and (4) we propose a new uncertainty-aware point-based few-shot object detector, taking into account the imperfection of pseudo GT bounding boxes.
    % The uncertainty estimation boosts the performance significantly over the baseline.
    % \item We propose two new relative metrics for object counting considering the practical situation of object counting. That is, these metrics penalize more when objects are wrongly counted on images containing small number of objects and penalize less otherwise. 

% \vspace{-10pt}

\section{Related Work}
\label{sec:related_work}

In this section, we review some related work on object counting and detection.

\myheading{Visual counting} focuses on some predefined classes such as car \cite{mundhenk2016large,Hsieh_2017_ICCV}, cell \cite{ARTETA20163,doi:10.1080/21681163.2016.1149104}, and human \cite{peng2018detecting,hu2020,Lian_2019_CVPR,fang2019locality,nwpu_2021,sindagi2019pushing,idrees2018composition,7780439,7298684,m_Ranjan-etal-ECCV18,m_Ranjan-etal-ACCV20,m_Abousamra-etal-AAAI21}. The methods can be grouped into two types: density-map-based and detection-based. The former, e.g., \cite{wang2020DMCount,Liu_2019_CVPR},  predicts and sums over density map from input image to get the final results. 
The latter (e.g., \cite{Hsieh_2017_ICCV,goldman2019precise}) counts the number of objects based on the detected boxes. The latter is better at justifying the counting number, however, it requires the GT bounding boxes for training and its performance is exceeded by that of the former, especially for images of crowded scenes.  

% These approaches are better at explaining the result but underperform in final counting results where most objects are not well separated. 
% 
\myheading{Few-shot counting (FSC)} counts the number of objects in an image with some exemplar bounding boxes of a new object class. Since the number of exemplar boxes is so small that an object detector cannot be reliably learned, prior methods on FSC are all based on density-map regression. GMN~\cite{Lu18} formulates object counting as object matching in video tracking such that a class agnostic counting module can be pretrained on a large-scale video object tracking dataset (ImageNet VID \cite{ILSVRC15}).
FamNet \cite{m_Ranjan-etal-CVPR21} correlates the features extracted from a few exemplars with the feature map to obtain the density map for object counting. VCN \cite{ranjan2022vicinal} improves upon \cite{m_Ranjan-etal-CVPR21} by augmenting the input image with different styles to make the counting more robust. LaoNet \cite{lin2021object} combines self-attention and cross attention in the transformer to aggregate features from the exemplar to the image to facilitate density map prediction. ICFR \cite{you2022iterative} proposes an iterative
framework to progressively refine the exemplar-related features, thus producing a better density map than a single correlation in \cite{m_Ranjan-etal-CVPR21}.
However, these approaches do not output object bounding boxes. An extension for object detection from these approaches is depicted in \Fref{fig:baseline}, but it has several limitations as illustrated \Fref{fig:common_errors_1} and \Fref{fig:common_errors_2}. Whereas, our approach \Approach~effectively predicts object bounding boxes along with the object count with only the supervision of FSC. 

\myheading{Object detection} methods include anchor-based approaches such as Faster-RCNN \cite{ren2015faster} and Retina Net \cite{lin2017focal}, point-based approaches such as like FCOS \cite{tian2019fcos} and Center-Net \cite{zhou2019objects}, and transformer-based approaches such as DETR \cite{carion2020end}, Point DETR \cite{chen2021points} and Anchor DETR \cite{wang2021anchor}. DETR is the first approach to apply transformer architecture \cite{vaswani2017attention} to object detection. Anchor DETR improves the convergence rate and performance of DETR by learnable anchor points representing the initial prediction of the objects in the image. 
% Point DETR also uses points as input to address the weakly learning with dot annotation. 
However, these methods require thousands of bounding box annotations on some predefined classes for training and cannot generalize well on a new class in testing with a few box exemplars as in our few-shot setting. 
Point DETR \cite{chen2021points} alleviates this requirement using two separate detectors: teacher (i.e., Point-DETR) and student (i.e., FCOS). The former learns from a small set of fully annotated boxes to generate pseudo-GT bounding boxes of a large amount of point-annotated images. Then the latter is trained with these pseudo-GT boxes to predict the bounding boxes of the test images. This approach is complicated and does not take into account the imperfect pseudo-GT bounding boxes.  
In contrast, our \Approach~is a unified single network with the uncertainty-aware bounding prediction. 

\myheading{Few-shot object detection (FSOD)} approaches are mostly based on Faster-RCNN \cite{ren2015faster} and can be divided into two subgroups based on episodic training \cite{kang2019few,yanICCV19metarcnn,zhang2021meta,xiao2020few,fan2020few} and fine-tuning  \cite{chen2018lstd,wang2020frustratingly,wu2020multi,fan2021generalized}. The former leverages episodic training technique to mimic the evaluation setting in the training whereas the latter fine-tunes some layers while keeping the rest unchanged to preserve the knowledge learned from the training classes. 
However, all FSOD approaches require box annotations for all objects of the base classes in training. It is not the case in our setting where only a few exemplar bounding boxes are given in training. To address this problem, we propose a two-stage training strategy wherein the first stage, the pseudo GT bounding boxes for all objects are generated from the given exemplar boxes and the dot annotations.
% However, in crowded scene like in the setting of FSCD, these approaches show inferior performance due to the highly occluded and dense objects of the same class packed in a small region. By contrast, our \Approach~treats objects as points to flexibly represent objects in any levels of visibility and density. 

% Although \Approach~is also a FSOD, it overcomes this limitation by using points instead of bounding boxes for representing objects. 

\myheading{Object detection with uncertainty} accounts for the uncertainty in the input image due to blurring or indistinguishable boundaries between objects and the background. Prior work \cite{he2019bounding,lee2020localization} assumes the object bounding boxes are characterized by a Gaussian distribution whose mean and standard deviation are predicted by a network trained with an uncertainty loss function derived from maximum the likelihood between the predicted distribution and the GT boxes. \cite{feng2021labels,feng2020labels} apply the uncertainty loss for 3D object detection. Our uncertainty loss shares some similarities with prior work, however, we use the uncertainty loss to account for the imperfection of the pseudo GT bounding boxes (not the input image) such that the Laplace distribution works significantly better than the prior Gaussian distribution as shown in experiments. 

% \vspace{-5pt}

\section{Proposed Approach}
\label{sec:approach}

\myheading{Problem definition:} In training, we are given a set of images containing multiple object categories. For each image $I$, a few exemplar bounding boxes $B_k$, $k=1,\dots,K$ where $K$ is the number of exemplars, and the dot annotations for all object instances of a target class are annotated. This kind of supervision is the same as in few-shot object counting. In testing, given a query image with bounding boxes for a few exemplar objects in the target class, our goal is to detect and count all instances of the target class in the query image. 

To address this problem, we propose a novel uncertainty-aware point-based few-shot object detector, named \Approach~trained with a novel two-stage training strategy to first generate pseudo GT bounding boxes from dot annotations and then train \Approach~on the generated pseudo GT boxes to predict bounding boxes of a new object class defined by a few bounding box exemplars in testing. 
The overview of \Approach~is illustrated in Fig.~\ref{fig:main_diagram}. 
% In the next subsections, we will discuss the feature extraction and feature aggregation to produce the exemplar-integrated feature map in Sec.~\ref{sec:fam}, then Sec.~\ref{sec:detr} specifies the transformer of \Approach, and Sec.~\ref{sec:two_stage} finally presents the two training stages.

\subsection{Feature Extraction and Feature Aggregation}
\label{sec:fam}

\myheading{Feature extraction:} A CNN backbone is used to extract feature map $F^I \in \mathbb{R}^{H \times W \times D}$ from the input image $I$ where $H, W, D$ are height, width, and number of channels of the feature map, respectively. We then extract the exemplar feature vectors $f^B_k \in \mathbb{R}^{1 \times D}$, at the center of the exemplar bounding boxes $B_k$. Finally, the exemplar feature vector $f^B \in \mathbb{R}^{1 \times D}$ is obtained by averaging these feature vectors, or $f^B = \frac{1}{K} \sum_{k} f^B_k$.

% \noindent 
\myheading{Feature aggregation:} We integrate the exemplar feature $f^B$ to the feature map of the image $F^I$ to produce the exemplar-integrated feature map $F^A$: 
\begin{equation}
    F^{A} =  W_{proj} * [F^{I}; F^{I} \otimes f^{B}],
\end{equation}
where $*, \otimes, [\cdot;\cdot]$ are the convolution, channel-wise multiplication, and concatenation operations, respectively. $W_{proj} \in \mathbb{R}^{2D \times D}$ is a linear projection weight. The first term in the concatenation preserves the original information of the feature map, while the second term aims at enhancing features at locations whose appearance are similar to those of the exemplars and suppressing the others.

\subsection{The Encoder-Decoder Transformer}
\label{sec:detr}

Inspired by DETR~\cite{carion2020end}, we design our transformer of \Approach~to take as input the exemplar-integrated feature map $F^A$ and $M$ query points $\{p_m\}_{m=1}^M$ and predict the bounding box $b_m$ for each query point $p_m$. The queries are the 2D points representing the initial guesses for the object locations rather than the learnable embeddings to achieve a faster training rate as shown in~\cite{wang2021anchor}. Thus, \Approach~is point-based approach that leverages the given dot annotations as the queries to predict the pseudo GT bounding boxes. Also, the transformer consists of two sub-networks: encoder and decoder. The former aims at enhancing features among the input set of features with the self-attention operation. The latter allows all the query points to interact with the enhanced features from the encoder with the cross-attention operation, thus capturing global information. 

Next, the decoder is used to: (1) predict the classification score $s$ representing the presence or absence of the object at a particular location, (2) regress the object's bounding box $\mu$ represented by the offset $x, y$ from the GT object center to the query point along with its size $w, h$. Following \cite{wang2021anchor}, first, the Hungarian algorithm is used to match each of the GT bounding boxes with its corresponding predicted bounding boxes. Then for each pair of matched GT and predicted bounding boxes, the focal loss~\cite{lin2017focal} and the combination of $L_1$ loss and GIoU loss~\cite{rezatofighi2019generalized} are used as training loss functions. In particular, at each query point, the following loss is computed:
\begin{equation}
L_{\text{DETR}} = \lambda_1 \text{Focal}(s, s^*) + \lambda_2 L_1(\mu, \mu^*) + \lambda_3 \text{GIoU}(\mu, \mu^*),
\label{eq:detr_loss}
\end{equation}
where $s^*, \mu^*$ are the GT class label and bounding box, respectively. $\lambda_1, \lambda_2, \lambda_3$ are the coefficients of focal, $L_1$, and GIoU loss functions, respectively.

Notably, \Approach~also estimates the uncertainty $\sigma$ when training under the supervision of the imperfect pseudo GT bounding boxes $\tilde{\mu}$. This uncertainty is used to regularize the learning of bounding box $\mu$ such that a lower loss is incurred at the prediction with high uncertainty. We propose to use the following uncertainty loss:
\begin{equation}
L_{\text{uncertainty}} = \frac{1}{2} \sum_{o \in \{x, y, w, h\}} {\frac{|\mu_o - \tilde{\mu}_o|}{\sigma_o} + \log \sigma_o},  
\label{eq:uncertainty_loss}
\end{equation}
where $\sigma$ is the estimated uncertainty. This loss is derived from the maximum likelihood estimation (MLE) between the predicted bounding box distribution characterized by a Laplace distribution and the pseudo GT bounding box as evidence. Another option is the Gaussian distribution, however, in the experiments, we show that the Gaussian has the inferior performance to that of Laplace, and is even worse than the variant that does not employ uncertainty estimation.
% \khoi{In qualitative results, show some examples of uncertainty predicted} 
% \Thanh: I drew but  objects are densely packed and small which very hard to see clearly.  I found that most of predicted b value in laplace distribution are closs to 99. which isn't valueable in visualize

\subsection{The Two-stage Training Strategy}
\label{sec:two_stage}
% \thanh{Introduce to PointDETR}
The proposed few-shot object detector, \Approach, can only be trained with the bounding box supervision for all objects. However, we only have bounding box annotation for a few exemplars and the point annotation for all objects as the setting of FSCD. Hence, we propose a two-stage training strategy as follows.

% To address this problem, Point DETR \cite{chen2021points} mitigate this problem by using two-stage training which learns the teacher model from a small fully annotated training set and then generates larger pseudo label to train student model. However, the quality of pseudo label from teacher model is not accounted when train student model.  

% \thanh{add uncertainty for our data} 
% We propose to integrate  the two-stage training strategy,  In our pipeline, the pseudo GT bounding boxes are generated from the dot annotations in the first stage to train the \Approach~in the second stage. The second stage mitigates the imperfection of pseudo label by uncertainty loss and uncertainty prediction module.

\begin{figure}[t]
\centering
\includegraphics[scale=0.18]{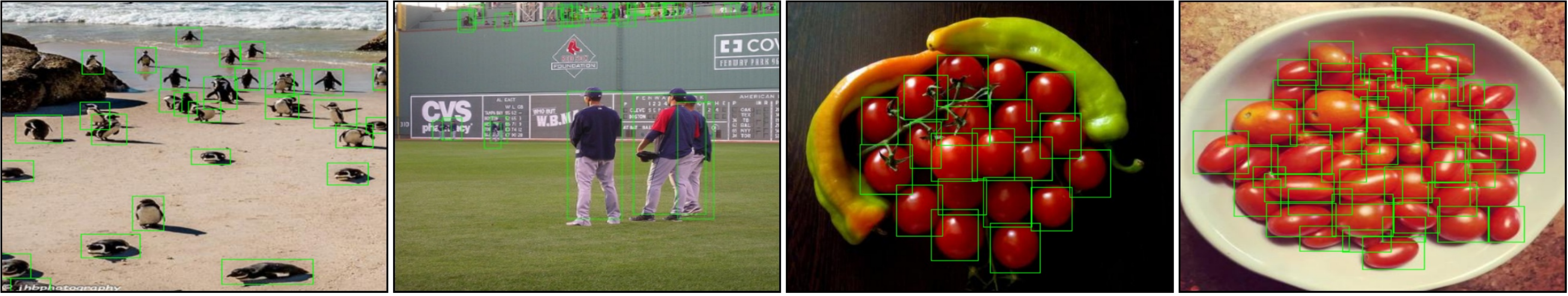}
\vspace{-10pt}
\caption{Examples of pseudo GT bounding boxes generated by the 1-st stage our method.}
\label{fig:pseudo-box}
\vspace{-10pt}
\end{figure}

In Stage 1, we first pretrain \Approach~on a few exemplar bounding boxes with their centers as the query points (as described in \Sref{sec:detr}). Subsequently, the pretrained network is used to predict the pseudo GT bounding boxes on the training images with the dot annotations as the query points. It is worth noting that, in this stage, we have the GT exemplar center as queries and their corresponding bounding boxes as supervision, so we do not use the Hungarian matching, uncertainty estimation, and uncertainty loss, i.e., we only use $L_{\text{DETR}}$ in Eq.~\eqref{eq:detr_loss} to train our \Approach. The visualization of some generated pseudo-GT boxes is illustrated in Fig.~\ref{fig:pseudo-box}.

In Stage 2, the generated pseudo GT bounding boxes on the training images are used to fine-tune the pretrained \Approach. The fine-tuned model is then used to make predictions on the test images with the uniformly sampled anchor points as queries. Different from Stage 1, the supervision is the imperfect pseudo GT bounding boxes, hence, we additionally leverage the uncertainty estimation branch with uncertainty loss to train. Particularly, we use the following loss to train \Approach~in this stage:
\begin{equation}
L_{\text{combine}} = L_\text{DETR} + \lambda_4 L_{\text{uncertainty}},
\label{eq:combine_loss}
\end{equation}
where $\lambda_4$ is the coefficient of $L_{\text{uncertainty}}$.

\begin{figure}[t]
  \subfloat[FSCD-147]{
	\begin{minipage}[c]{
	   0.49\textwidth}
	   \centering
	   \label{fig:fscd-147}
	   \includegraphics[width=1\textwidth]{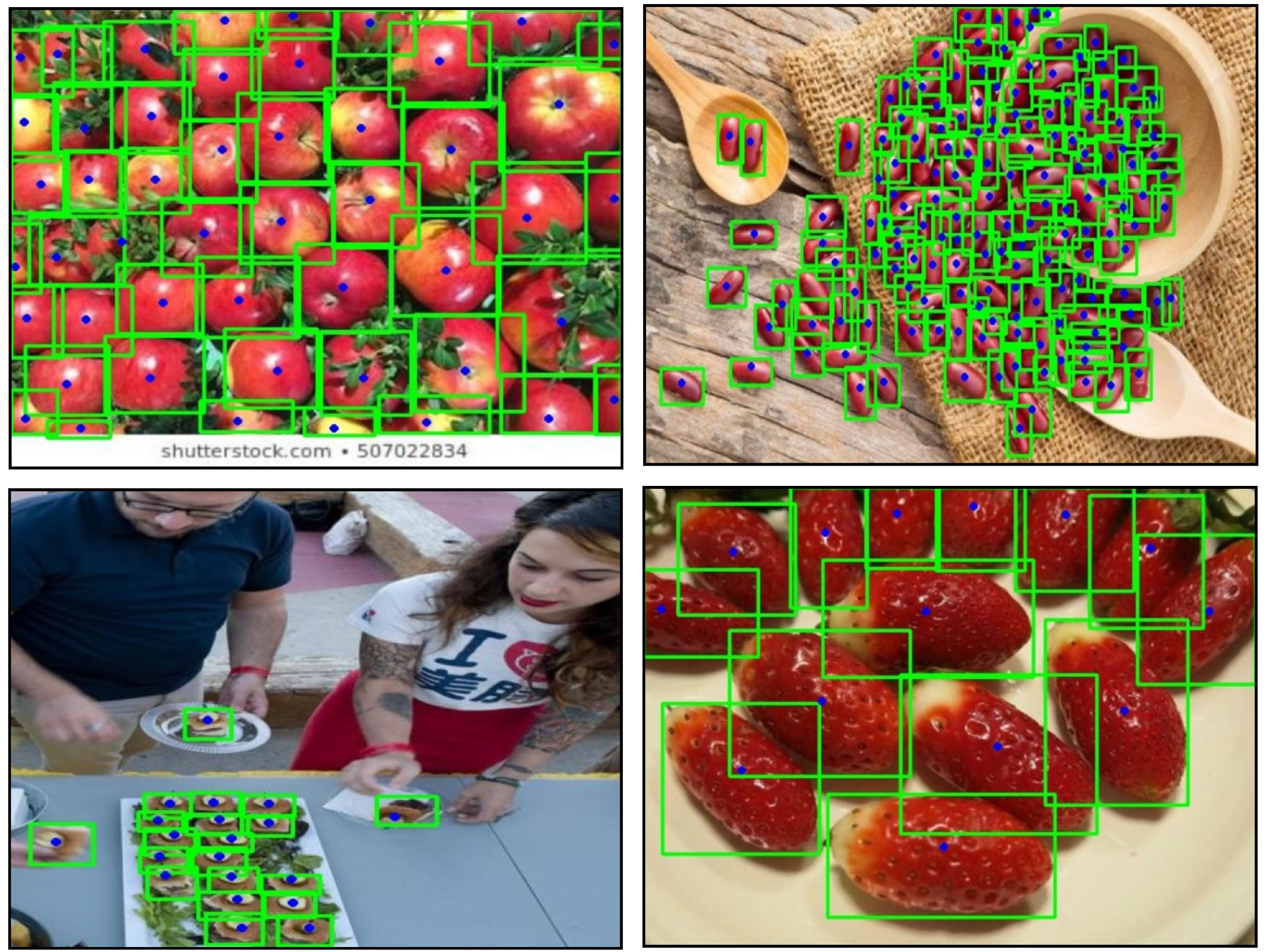}
	\end{minipage}}
  \subfloat[FSCD-LVIS]{
	\begin{minipage}[c]{
	   0.49\textwidth}
	   \centering
	   \label{fig:fscd-lvis}
	   \includegraphics[width=\textwidth]{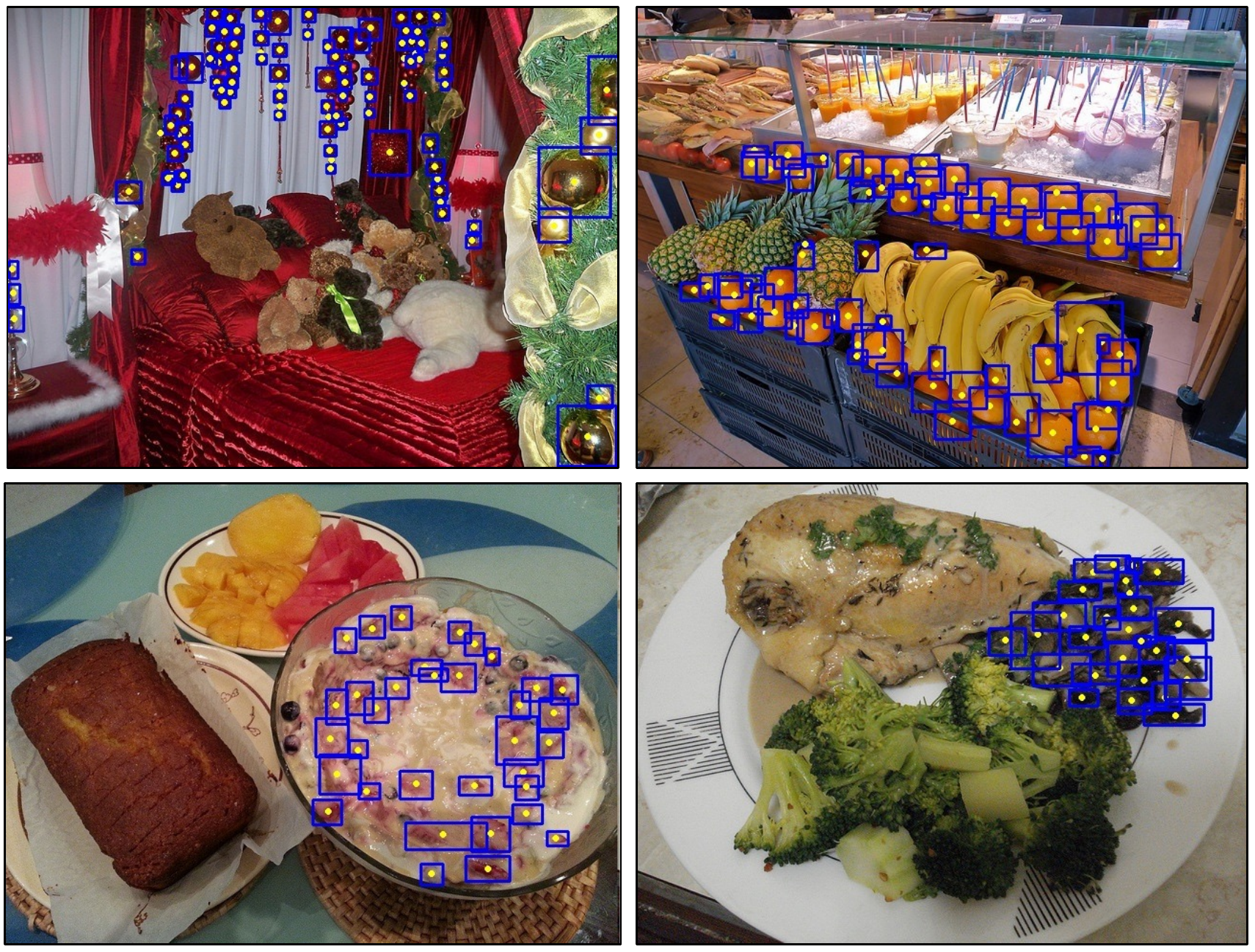}
	\end{minipage}}
%   \subfloat[]{
% 	\begin{minipage}[c]{
% 	   0.22\textwidth}
% 	   \centering
% 	   \label{fig:fscd-lvis}
% 	   \includegraphics[width=\textwidth]{images/LVIS_stat.pdf}
% 	\end{minipage}}
\vspace{-5pt}
\caption{Sample images from our datasets and annotated bounding boxes.
% ; (c) Histogram of \# objects per image of the FSCD-LVIS dataset
}
\vspace{-10pt}
\end{figure}

\setlength{\tabcolsep}{10pt}
\begin{table}[t]
\centering
	\caption{Comparison between the FSCD-147 and FSCD-LVIS datasets 
% 	in \# classes and \# total, train, val, test images
	}
	\vskip -0.1in
	\label{tab:comparison}
	\centering
	\begin{tabular}{lccccc}
    \toprule 
    & & \multicolumn{4}{c}{Number of images} \\
    \cmidrule(lr){3-6}
    Dataset & \#classes & total & train & val & test \\
    \midrule 
    FSCD-147 & 147 & 6135 & 3659 & 1286 & 1190 \\
    FSCD-LVIS & 372 & 6195 & 4000 & 1181 & 1014 \\
    % FSCD-LVIS unseen & 372 & 6195 & 3953  & 0 & 2242 \\
    \bottomrule 
    \end{tabular}
    % \vspace{-10pt}
\end{table}

% [2628, 1212, 684, 402, 286, 959]

\begin{table}[t]
\centering
\setlength{\tabcolsep}{5pt}
	\caption{Number of images for each bin of the FSCD-LVIS dataset}
	\label{tab:histogram}
	\vskip -0.1in
	\centering
	\begin{tabular}{lcccccc}
    \toprule 
    Counting range & 20-30 & 30-40 & 40-50 & 50-60 & 60-70 & $>$70 \\ 
    \midrule 
    \# images in range & 2628 & 1212 & 684 & 402 & 286 & 959 \\
    \bottomrule 
    \end{tabular}
    \vspace{-10pt}
\end{table}

\section{New Datasets for Few-shot Counting and Detection}
\label{sec:datasets}

A contribution of our paper is the introduction of two new datasets for few-shot counting and detection. In this section, we will describe these datasets.  

\subsection{The FSCD-147 Dataset}

The FSC-147 dataset \cite{m_Ranjan-etal-CVPR21} was recently introduced for the few-shot object counting task with 6135 images across a diverse set of 147 object categories.
% The train, validation, and test sets consist of 3659, 1286, and 1190 images, respectively. 
In each image, the dot annotation for all objects and three exemplar boxes are provided. However, this dataset does not contain bounding box annotations for all objects. For evaluation purposes, we extend the FSC-147 dataset by providing bounding box annotations for all objects of the val and test sets. We name the new dataset FSC\textbf{D}-147.
To be consistent with the counting, an object will be annotated with its bounding box only if it has a dot annotation. Fig.~\ref{fig:fscd-147} shows some samples of the FSCD-147 dataset. %For further detail of the dataset, please refer \cite{m_Ranjan-etal-CVPR21}. 
It is worth noting that annotating bounding boxes for many objects in crowded scenes of  FSC-147 is a laborious process, and this is a significant contribution of our paper. 

%all the images of FSC-147 is targeted for counting objects in the crowded scene, thus annotating these bounding boxes is a difficult job, especially with those highly dense images where it is hard to recognize the clear boundary between objects. 

\subsection{The FSCD-LVIS Dataset}

Although the FSC-147 dataset contains images with a large number of objects in each image, the scene of each image is rather simple. Each image of FSC-147 shows the target object class so clearly that one can easily know which object class to count without having to specify any exemplars as shown in Fig.~\ref{fig:problem_statement} and Fig.~\ref{fig:fscd-147}. For real-world deployment of methods for few-shot counting and detection, we introduce a new dataset called FSCD-LVIS. Specifically, the scene is more complex with multiple object classes with multiple object instances each as illustrated in Fig.~\ref{fig:fscd-lvis}. Without providing the exemplars for the target class, one cannot definitely guess which the target class is. 

The FSCD-LVIS dataset contains 6196 images and 377 classes, extracted from the LVIS dataset \cite{gupta2019lvis}. For each image, we filter out all instances with an area smaller than 20 pixels, or a width or a height smaller than 4 pixels. The comparison between the FSCD-LVIS and FSC-147 datasets is shown in Tab.~\ref{tab:comparison}. 
The histogram of the number of labeled objects per image is illustrated in Tab.~\ref{tab:histogram}. The LVIS dataset has the box annotations for all objects, however, to be consistent with the setting of FSCD, we randomly choose three annotated bounding boxes of a selected object class as the exemplars for each image in the training set of FSCD-LVIS.

\section{Experimental Results}
\label{sec:experiments}

% \textbf{Datasets:} We evaluate our approach the FSCD-147 and the FSCD-LVIS datasets as described in Sec.~\ref{sec:datasets}

\textbf{Metrics.} 
For object counting, we use Mean Average Error (MAE) and Root Mean Squared Error (RMSE), which are standard measures used in the counting literature. Besides, the Normalized Relative Error (NAE) and Squared Relative Error (SRE) are also adopted. In particular, 
$
    \text{MAE} = \frac{1}{J} \sum_{j=1}^{J} |c^*_{j} - c_{j}|; 
    \text{RMSE} = \sqrt{\frac{1}{J} \sum_{j=1}^{J} ({c^*_{j} - c_{j}})^{2}}; 
    \text{NAE} = \frac{1}{J} \sum_{j=1}^{J} \frac{|c^*_{j} - c_{j}|}{c^*_{j}}; 
    \text{SRE} = \sqrt{\frac{1}{J} \sum_{j=1}^{J} \frac{({c^*_{j} - c_{j}})^{2}}{c^*_{j}} }
$
% 
% {
% \footnotesize
% \begin{align}
%     \text{MAE}  &= \frac{1}{J} \sum_{j=1}^{J} \norm{c^*_{j} - c_{j}}; &
%     \text{RMSE} = \sqrt{\frac{1}{J} \sum_{j=1}^{J} ({c^*_{j} - c_{j}})^{2}}; \\
%     \text{NAE} &= \frac{1}{J} \sum_{j=1}^{J} \frac{\norm{c^*_{j} - c_{j}}}{c^*_{j}} ; &
%     \text{SRE} = \sqrt{\frac{1}{J} \sum_{j=1}^{J} \frac{({c^*_{j} - c_{j}})^{2}}{c^*_{j}} };
% \end{align}
% }% 
% {
% \tiny
% \begin{align}
%     \text{MAE} = \frac{1}{J} \sum_{j=1}^{J} \norm{c^*_{j} - c_{j}}; 
%     \text{RMSE} = \sqrt{\frac{1}{J} \sum_{j=1}^{J} ({c^*_{j} - c_{j}})^{2}}; 
%     \text{NAE} = \frac{1}{J} \sum_{j=1}^{J} \frac{\norm{c^*_{j} - c_{j}}}{c^*_{j}}; 
%     \text{SRE} = \sqrt{\frac{1}{J} \sum_{j=1}^{J} \frac{({c^*_{j} - c_{j}})^{2}}{c^*_{j}} } \nonumber
% \end{align}
% }% 
where $J$ is the number of test images, $c^*_{j}$ and $c_{j}$ are GT and the predicted number of objects for image $j$, respectively. Unlike the absolute errors MAE and RMSE, the relative errors NAE and SRE reflect the practical usage of visual counting, i.e., with the same number of wrong objects counted (e.g., 10), it is more serious for images having a smaller number of objects (e.g., 20) than the ones having larger numbers of objects (e.g., 200).

For object detection, we use mAP and AP50. They are the average precision metrics with the IoU threshold between predicted and GT boxes for determining a correct prediction ranging from 0.5 to 0.95 for mAP and 0.5 for AP50.

\myheading{Implementation details.}
We implement our approach, baselines, and ablations in PyTorch \cite{NEURIPS2019_9015}. Our backbone network is ResNet-50 \cite{he2016deep} with the frozen Batch Norm layer \cite{ioffe2015batch}.
We extract exemplar features $f^B_k$ from exemplar boxes $B_k$ from Layer 4 of the backbone. Our transformer network shares the same architecture as that of Anchor DETR \cite{wang2021anchor} with the new uncertainty estimation and is trained with additional uncertainty loss as described in Sec.~\ref{sec:detr}, while keeping the rest intact with six layers for both encoder and decoder. We use AdamW optimizer \cite{loshchilov2017decoupled} with the learning rate of $10^{-5}$ for the backbone and $10^{-4}$ for the transformer to train \Approach~in 30 epochs with a batch size of one. We use the following training loss coefficients $\lambda_1=2, \lambda_2=5, \lambda_3=2, \lambda_4=2$, which were tuned based on the validation set. Also, the number of exemplar boxes is set to $K=3$, as in FamNet \cite{m_Ranjan-etal-CVPR21} for a fair comparison.

% \begin{figure}[t]
%     \centering
%     \includegraphics[width=8cm]{images/Baseline.drawio.pdf}
%     \caption{}
%     \label{fig:ridge_baseline}
% \end{figure}
\subsection{Ablation Study}

We conduct several experiments on the validation data of FSCD-147 to study the contribution of various components of our method. 
% the contribution of each component to the final performance gain of \Approach~as well as the sensitivity analysis of some important hyper-parameters 

% (1) how each component contribute to final result: feature aggregation module, transformer, (2) how performance changes with different type of anchor points, (3) how performance changes when number of query pattern per anchor point change, (4) how performance changes when number of anchor points vary, (4)   (2) how performance changes when we vary number of exemplar boxes (2) how performance changes when we vary number of exemplar boxes

\setlength{\tabcolsep}{3pt}
\begin{table}[t]
    \small
    \centering
    \caption{Ablation study on each component's contribution to the final results}
    \vskip -0.1in 
    \begin{tabular}{c c | c c c c  c c} 
        \toprule 
        \multicolumn{2}{c}{Combination} & \multicolumn{4}{c}{Counting} & \multicolumn{2}{c}{Detection} \\
        \cmidrule(lr){1-2} \cmidrule(lr){3-6} \cmidrule(lr){7-8}
         Pseudo box & Uncertainty  & MAE ($\downarrow$) & RMSE($\downarrow$) & NAE($\downarrow$) & SRE ($\downarrow$) & AP($\uparrow$) & AP50($\uparrow$)  \\ 
        \midrule 
        \checkmark & \checkmark & \textbf{20.38} & \textbf{82.45} & \textbf{0.19} & \textbf{3.38} & \textbf{17.27} & \textbf{41.90}  \\
        \xmark & \checkmark & 29.74 & 104.04 & 0.26 & 4.44 & 11.37 & 29.98 \\
        \checkmark & \xmark & 23.57 & 93.54 & 0.21 & 3.77 & 14.19 & 36.34   \\
        \xmark &\xmark  & 31.36 & 105.76 & 0.27 & 4.60 & 10.81 & 28.76 \\
        % FSDetView+Our-1st &  39.89  & 123.58 & 0.56 & 5.85 & 13.57 & 28.24\\
        % FSDetView+RR  & 41.58 & 125.42 & 0.65 & 6.08 & 10.20 & 26.46\\
        \bottomrule 
    \end{tabular}
    \label{tab:ablation} 
    % \vspace{-10pt}
\end{table}

\begin{table}[t]
    \small
    \centering
    \caption{Performance of \Approach~with different types of anchor points}
    \vskip -0.01in
    \begin{tabular}{l c c c c  c c} 
        \toprule 
         & \multicolumn{4}{c}{Counting} & \multicolumn{2}{c}{Detection} \\    
          \cmidrule(lr){2-5} \cmidrule(lr){6-7}
        Anchor type & MAE ($\downarrow$) & RMSE($\downarrow$) & NAE($\downarrow$) & SRE ($\downarrow$) & AP($\uparrow$) & AP50($\uparrow$) \\ 
        \midrule 
        Learnable & 25.20 & \textbf{81.94} & 0.25 & 3.92 & 16.46 & 38.34  \\ 
        Fixed grid (proposed) & \textbf{20.38} & 82.45 & \textbf{0.19}& \textbf{3.38} & \textbf{17.27} & \textbf{41.90} \\
        \bottomrule 
        %  \hline
    \end{tabular}
    \label{tab:Anchor_points_type} 
    %\vspace{-10pt}
\end{table}

% We have the following variants: 
% \begin{enumerate}
%     \item \Approach: our full approach
%     \item \Approach-uncertainty: remove the uncertainty estimation in the stage 2 of the training
%     \item \Approach+RR box: use the pseudo GT boxes generated from the FamNet+RR (as described in Fig.~\ref{fig:baseline}) with features extracted from dot annotations 
%     \item \Approach-uncert.+RR: remove the uncertainty estimation and use pseudo GT box generated from FamNet+RR
%     % \item FSDetView+Our-1st: train FSDetView with the pseudo GT boxes generated from the training stage 1 of \Approach 
%     % \item FSDetView+RR box: train FSDetView with the pseudo GT boxes generated from the FamNet+RR, similar to \Approach+RR
% \end{enumerate}

\myheading{Pseudo Box and Uncertainty Loss}. 
From Tab.~\ref{tab:ablation}, we see that pseudo GT boxes (pseudo box) generated from our first stage are much better than the boxes generated by Ridge Regression (-6 in AP, +9 in MAE). Without using uncertainty loss (similar to Point DETR \cite{chen2021points}), the performance drops substantially (-3 in AP, +3 in MAE). That justifies the effectiveness of our uncertainty loss. Without using both of them, the performance gets worst (-7 in AP, +11 in MAE). These results demonstrate the important contribution of our proposed pseudo GT box generation and uncertainty loss.

\myheading{Types of anchor points.} 
As described in Sec.~\ref{sec:detr}, we follow the design of Anchor DETR whose anchor points can either be learnable or fixed-grid. The results of these two types are shown in \Tref{tab:Anchor_points_type}, we can see that the fixed-grid anchor points are comparable to the learnable anchor points on counting metrics, but better on the detection metrics. Thus, the fixed-grid anchor points are chosen for the \Approach.

\myheading{Numbers of anchor points.} Tab.~\ref{tab:Query_points} presents the results with different numbers of anchor points $M$. Both the detection and counting results increase as the number of anchor points increases, and they reach the highest points when the number of anchor points is $M=600$. Hence, we choose 600 anchor points for \Approach.

\myheading{Types for the uncertainty loss.} Instead of using the Laplace distribution as described in Sec.~\ref{sec:detr}, we use Gaussian distribution to derive the uncertainty loss: $
L_{\text{uncertainty}}^{\text{Gaussian}} = \frac{1}{2} \sum_{o \in \{x, y, w, h\}} {\frac{(\mu_o - \mu_o^*)^2}{\sigma_o^2} + \log \sigma_o^2}.  
\label{eq:uncertainty_loss_gaussian}
$ This loss is similar to \cite{he2019bounding}. The results are shown in Tab.~\ref{tab:distributions}. The uncertainty loss derived from the Gaussian distribution yields the worst results among the variants, even worse than the variant without using any uncertainty loss. On the contrary, our proposed uncertainty loss derived from the Laplace distribution gives the best results. 

% generated from RR. both when with and without using uncertainty loss.  Our uncertainty loss also improve both counting and detection results over both fewshot baseline.  
% \khoi{please add your observation and comments.}
% 

\begin{table}[t]
    \centering
    \caption{Performance of \Approach~with different numbers of anchor points}
    \vspace{-5pt}
    \begin{tabular}{c c c c c  c c} 
         \toprule 
         & \multicolumn{4}{c}{Counting} & \multicolumn{2}{c}{Detection} \\    
          \cmidrule(lr){2-5} \cmidrule(lr){6-7}
         \# anchor points & MAE ($\downarrow$) & RMSE($\downarrow$) & NAE($\downarrow$) & SRE ($\downarrow$) & AP($\uparrow$) & AP50($\uparrow$)  \\ 
         \midrule 
         100 & 30.22 & 113.24 & 0.24 & 4.55 & 11.26 & 28.62 \\
         200 & 26.76 & 103.11 & 0.22 & 4.15 & 14.06 & 34.33 \\
         300 & 23.57 & 93.54 & 0.21 & 3.77 & 14.19 & 36.34 \\
         400 & 22.62 & 88.45 & 0.21 & 3.69 & 14.91 & 37.30 \\
         500 & 21.72 & 85.20 & 0.20 & 3.52 & 16.03 & 39.66 \\
        %  600 & 20.03 & 80.23 & 0.19 & 3.33  & 16.04 & 40.67 \\
         600 & \textbf{20.38} &	\textbf{82.45} & \textbf{0.19} & \textbf{3.38} & \textbf{17.27} & \textbf{41.90} \\
         700 & 21.19 & 83.70 & 0.22 & 3.47 & 15.10 & 37.85 \\
         \bottomrule 
        %  7 & 21.44 & 83.15 & 20.03 & ? & 16.03 & 39.93 \\
        % \hline
        % 8 & 22.35 & 90.55 & 20.43 & ? & 16.85 & 40.93 \\
        % \hline
        % 9 & 24.34 & 93.22 & 21.38 & ? & 16.75 & 40.37 \\
        % \hline
    \end{tabular}
    \label{tab:Query_points} 
    % \vspace{-10pt}
\end{table}

\begin{table}[t]
    \centering
    \caption{Ablation study for the  uncertainty loss}
    \vspace{-5pt}
    \begin{tabular}{l c c c c c c} 
         \toprule 
         & \multicolumn{4}{c}{Counting} & \multicolumn{2}{c}{Detection} \\    
          \cmidrule(lr){2-5} \cmidrule(lr){6-7}
         Distribution type & MAE ($\downarrow$) & RMSE($\downarrow$) & NAE($\downarrow$) & SRE ($\downarrow$) & AP($\uparrow$) & AP50($\uparrow$)  \\ 
         \midrule 
         W/o uncertainty loss & 23.20 & 92.87 & 0.21 &3.77 & 13.86 & 35.67 \\
         Gaussian loss & 24.46 & 94.20 & 0.22 & 3.84 & 14.03 & 34.91 \\ 
        %  \Approach & 22.80 & 91.84 & 0.20 & & 15.55 & 37.65 \\
         Laplacian loss (proposed) & \textbf{20.38} &	\textbf{82.45}&	\textbf{0.19}& \textbf{3.38} & \textbf{17.27} &\textbf{ 41.90} \\
         \bottomrule 
    \end{tabular}
    \label{tab:distributions} 
    % \vspace{-10pt}
\end{table}

% \begin{figure}[t]
% \centering
% \includegraphics[width=1\textwidth]{images/overlap_distribution.pdf}
% \vspace{-10pt}
% \caption{Overlap between train and val sets with Gaussian and Laplace distributions}
% \label{fig:overlap_distribution}
% \vspace{-10pt}
% \end{figure}

% \begin{figure}[t]
%   \subfloat[Absolute Errorsf]{
% 	\begin{minipage}[c]{
% 	   0.47\textwidth}
% 	   \centering
% 	   \label{fig:abs_errors}
% 	   \includegraphics[width=1\textwidth]{images/abso_diff.pdf}
% 	\end{minipage}}
%   \subfloat[Relative Errors]{
% 	\begin{minipage}[c]{
% 	   0.51\textwidth}
% 	   \centering
% 	   \label{fig:rel_errors}
% 	   \includegraphics[width=\textwidth]{images/rel_diff.pdf}
% 	\end{minipage}}

% \vspace{-5pt}
% \caption{Sample images from our datasets and annotated bounding boxes.
% % ; (c) Histogram of \# objects per image of the FSCD-LVIS dataset
% }
% \vspace{-10pt}
% \label{fig:compare_abs_n_rel_error_fscd_famnet}
% \end{figure}

\setlength{\tabcolsep}{2pt}

\begin{table}[t]
\small
\setlength{\tabcolsep}{1pt}
\centering
    \caption{Comparison with strong baselines on the FSCD-147 test set}
    \vskip -0.1in
    \begin{tabular}{l  c c c c  c c} 
\toprule 
%           & \multicolumn{6}{c}{FSCD-147} \\
%         \hline
         & \multicolumn{4}{c}{Counting} & \multicolumn{2}{c}{Detection} \\    
          \cmidrule(lr){2-5} \cmidrule(lr){6-7}
         Method & MAE ($\downarrow$) & RMSE ($\downarrow$) & NAE($\downarrow$) & SRE ($\downarrow$) & AP($\uparrow$) & AP50($\uparrow$)  \\ 
         \midrule 
         FamNet \cite{m_Ranjan-etal-CVPR21}+RR  & 22.09 & \textbf{99.55} & 0.44 & 6.45 & 9.44 & 29.73 \\ 
         FamNet \cite{m_Ranjan-etal-CVPR21}+MLP  & 22.09 & \textbf{99.55} & 0.44 & 6.45 & 1.21 & 6.12 \\ 
         Attention-RPN \cite{fan2020few}+RR box & 32.70 & 141.07 & 0.38 & 5.27 & 18.53 & 35.87 \\
         FSDetView \cite{xiao2020few}+RR box & 37.83 & 146.56  & 0.48 & 5.47 & 13.41 & 32.99 \\
        %  \Approach+RR box & 23.41 & 136.23 & 0.22 & 4.06 & 16.49 & 40.88  \\
        %  \Approach+RR box & 19.55 & 133.55 & 0.18 & 3.93 & 17.55 & 43.68 \\
        %  FamNet \cite{m_Ranjan-etal-CVPR21}+mean box  &  \\ 
        %  Attention-RPN \cite{fan2020few}+mean box  & 34.39 & 143.73 & 0.40 & 5.02 & 15.64 & 32.95\\
        %  FSDetView \cite{xiao2020few}+mean box  &  \\
        %  \hline
         Attention-RPN \cite{fan2020few}+pseudo box  & 32.42 &141.55 &0.38 &5.25 &20.97 &37.19\\
         FSDetView \cite{xiao2020few}+pseudo box  & 37.54 & 147.07 & 0.44 & 5.40 & 17.21 & 33.70 \\
         \Approach~(proposed)& \textbf{16.79} & 123.56  & \textbf{0.19} & \textbf{5.23} & \textbf{22.66} & \textbf{50.57} \\
        %  \hline \hline
        %  Upper bound & ? & ?  & ? & ? & ? & ? \\
         \bottomrule 
    \end{tabular}
    \label{tab:FSCD_results} 
\end{table}
\vspace{-10pt}

% 96.86	117.8	19.37	52.68	0.84	2.04
\begin{table}[t]
\small
\setlength{\tabcolsep}{1pt}
\centering
    \caption{Comparison with strong baselines on the FSCD-LVIS test set 
    % \Approach~outperforms all other methods by a large margin. Recall that \Approach~is trained with pseudo GT.
    % ; if the actual GT is used, this method achieves MAE, RMSE, NAE, SRE, AP, and AP50 of 18.51, 24.48, 0.45, 3.99, 4.92, and 14.49, respectively
    }
    % 17.63, 24.74, 3.53, 11.06, 6.54, and 17.00,
    \vskip -0.1in
    \begin{tabular}{l  c c c c  c c} 
         \toprule 
                  & \multicolumn{4}{c}{Counting} & \multicolumn{2}{c}{Detection} \\    
          \cmidrule(lr){2-5} \cmidrule(lr){6-7}
         Method & MAE ($\downarrow$) & RMSE ($\downarrow$) & NAE($\downarrow$) & SRE ($\downarrow$) & AP($\uparrow$) & AP50($\uparrow$)  \\ 
         \midrule 

         FamNet \cite{m_Ranjan-etal-CVPR21}+RR & 60.53 & 84.00 & 1.82 & 14.58 & 0.84 & 2.04\\ 
         Attention-RPN \cite{fan2020few}+RR box  & 61.31 & 64.10 & 1.02 & 6.94 & 3.28 & 9.44 \\
         FSDetView \cite{xiao2020few}+RR box & 26.81 & 33.18  & 0.56 & 4.51 & 1.96 & 6.70 \\
        %  \Approach+RR box & 13.19 & 18.42 & 2.64 & 8.24 & 2.21 & 6.62 \\
        %  Attention-RPN+Our-1st \cite{fan2020few} &  &  &  &  &  &  \\
        %  FSDetView+Our-1st \cite{xiao2020few} &  &   & ? & ? &  &  \\
        %  FamNet \cite{m_Ranjan-etal-CVPR21}+mean box  &  \\ 
        %  Attention-RPN \cite{fan2020few}+mean box  & \\
        %  FSDetView \cite{xiao2020few}+mean box  &  \\
        %  \hline
         Attention-RPN \cite{fan2020few}+pseudo box  & 62.13 &65.16 &1.07 &7.21 &4.08 &11.15\\
         FSDetView \cite{xiao2020few}+pseudo box  &  24.89 &31.34 &0.54 &4.46 &2.72 &7.57\\
        %  \Approach & \textbf{18.51} & \textbf{24.48} & \textbf{0.45} & \textbf{4.24} & \textbf{4.92} & \textbf{14.49} \\
          \Approach & \textbf{18.51} & \textbf{24.48} & \textbf{0.45} & \textbf{3.99} & \textbf{4.92} & \textbf{14.49} \\
%         \hline
%         \hline 
        %  Upper-bound & 15.5 & 26.78& 3.1 & 11.98 & 3.91 & 11.78 \\
        %  \hline
        % Thanh: lan trc khi em eval, em de score=1 de debug, nen ket qua kem. Em update lai score = pred_score dc ket qua nhu, tren. 
%         \Approach+GT box & 17.63 & 24.74 &  3.53 & 11.06 & 6.54 & 17.00 \\
         \bottomrule 
    \end{tabular}
    \label{tab:LVIS_results} 
    % \vspace{-10pt}
\end{table}

% 96.86	117.8	19.37	52.68	0.84	2.04
\begin{table}[t]
\small
\setlength{\tabcolsep}{1pt}
\centering
    \caption{Comparison on FSCD-LVIS with unseen test classes}
    \vskip -0.1in
    \begin{tabular}{l  c c c c  c c} 
         \toprule 
                  & \multicolumn{4}{c}{Counting} & \multicolumn{2}{c}{Detection} \\    
          \cmidrule(lr){2-5} \cmidrule(lr){6-7}
         Method & MAE ($\downarrow$) & RMSE ($\downarrow$) & NAE($\downarrow$) & SRE ($\downarrow$) & AP($\uparrow$) & AP50($\uparrow$)  \\ 
         \midrule 
        %  FamNet \cite{m_Ranjan-etal-CVPR21}+RR & 102.96 & 125.93 & 20.59 & 56.32 & 0.64 & 1.97\\ 
         
         FamNet \cite{m_Ranjan-etal-CVPR21}+RR & 68.45 & 93.31 & 2.34 & 17.41 & 0.07 & 0.30\\ 
         Attention-RPN \cite{fan2020few}+RR box & 35.55 & 42.82 & 1.21 & 7.47 & 2.52 & 7.86 \\
         FSDetView \cite{xiao2020few}+RR box & 28.56 & 39.72 & 0.73 & 4.88 & 0.89 & 2.38 \\
         Attention-RPN \cite{fan2020few}+pseudo box  & 39.16 & 46.09 & 1.34 & 8.18 & 3.15 & 7.87\\
         FSDetView \cite{xiao2020few}+pseudo box  & 28.99 & 40.08 & 0.75 & 4.93 & 1.03 & 2.89 \\
        %  \Approach & \textbf{18.51} & \textbf{24.48} & \textbf{0.45} & \textbf{4.24} & \textbf{4.92} & \textbf{14.49} \\
          \Approach & \textbf{23.50} & \textbf{35.89} & \textbf{0.57} & \textbf{4.17} & \textbf{3.85} & \textbf{11.28} \\
%         \hline
%         \hline 
        %  Upper-bound & 15.5 & 26.78& 3.1 & 11.98 & 3.91 & 11.78 \\
        %  \hline
        % Thanh: lan trc khi em eval, em de score=1 de debug, nen ket qua kem. Em update lai score = pred_score dc ket qua nhu, tren. 
%         \Approach+GT box & 17.63 & 24.74 &  3.53 & 11.06 & 6.54 & 17.00 \\
         \bottomrule 
    \end{tabular}
    \label{tab:LVIS_results_unseen} 
    \vspace{-10pt}
\end{table}

\begin{figure}[t!]
\centering
\includegraphics[width=1\textwidth]{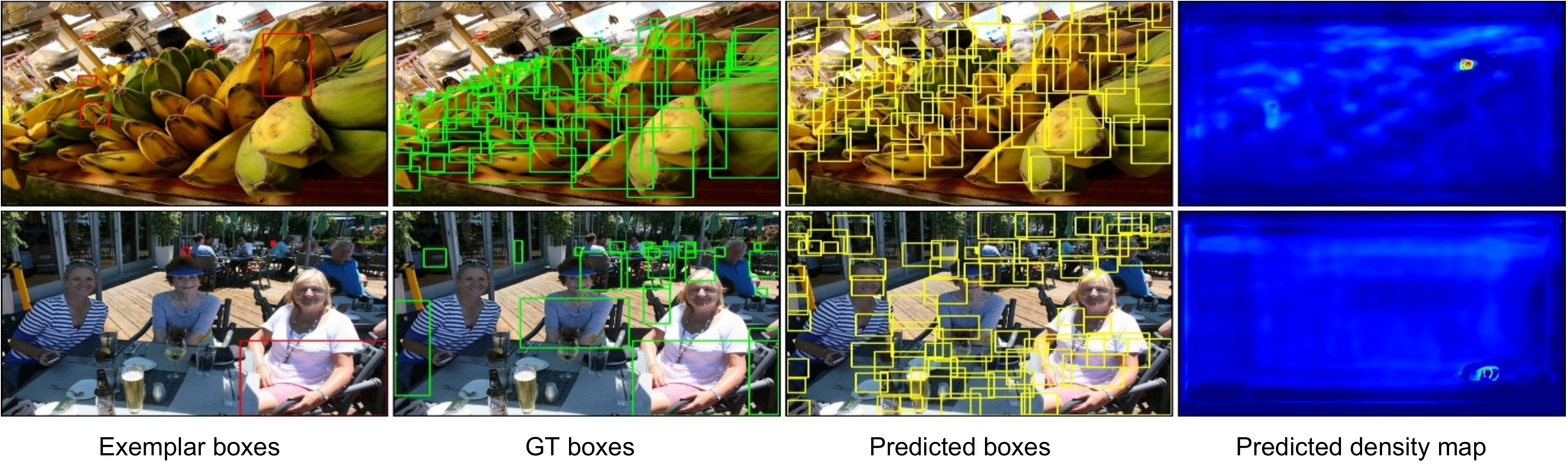}
\vspace{-20pt}
\caption{Results of FamNet on the FSCD-LVIS dataset. The objects of interest for Row 1 and 2 are bananas and chairs, respectively. FamNet fails to output good density maps for images containing objects with huge variations in shape and size.}
% \mhoai{If I am a reader who just read the captions of the figures, I don't know what this figure says. Expand the caption to tell the message. What's the message here? Is the message FamNet not good enough here? }}
\label{fig:famnet_on_lvis}
% \vspace{-10pt}
\end{figure}

% 
% Fig.~\ref{fig:overlap_distribution} investigates the two distributions with the zero mean and average standard deviation of all predicted bounding boxes from the train and test sets. The Laplace distribution shows better overlap between train and test than that of the Gaussian. This justifies our choice of distribution for deriving the uncertainty loss. \mhoai{I don't understand this argument and the figure 6} \khoi{we assume the results are better when the distribution of object width in the train set is highly overlapped (fits) with the distribution of object width in the test set}

\subsection{Comparison to Prior Work}

Since there is no existing method for the new FSCD task, we compare \Approach~with several strong baselines adapted from few-shot object counting and few-shot object detection: FamNet \cite{m_Ranjan-etal-CVPR21}+RR, FamNet \cite{m_Ranjan-etal-CVPR21}+MLP, Attention-RPN \cite{fan2020few}+RR, and FSDetView \cite{xiao2020few}+RR. Other few-shot object detectors are not chosen due to the unavailability of the source code or the requirement for fine-tuning on the whole novel classes together (see Sec.~\ref{sec:related_work}). It is different from our setting, where each novel class is processed independently in a separate image.
FamNet+RR is a method that uses Ridge Regression on top of the density map predicted by FamNet as depicted in Fig.~\ref{fig:baseline}. FamNet+MLP is similar to FamNet+RR but replaces the ridge regression with a two-layer MLP with the Layer norm.
% and uses the Non-maximum suppression (NMS) to eliminate highly overlapped predicted boxes to obtain the final results.
Attention-RPN and FSDetView are detection-based methods, which require GT bounding boxes for all objects to train, thus, we generate the pseudo GT bounding boxes using either (1) the FamNet+RR with the features extracted from the dot annotations of training images instead of peak locations (called RR box) or (2) our first stage of training as described in Sec.~\ref{sec:two_stage} (called pseudo box).
Tab.~\ref{tab:FSCD_results} and Tab.~\ref{tab:LVIS_results} show the comparison on the test sets of FSCD-147 and FSCD-LVIS, respectively. 
% Attention-RPN is based on Faster-RCNN \cite{ren2015faster} with multiple levels of aggregation between the support and query images including attention RPN, multi-relation detector.
% FSDetView uses a new feature aggregation module comprising identity, channel-wise multiplication, and subtraction operations between support and query feature map.
% 
% It is worth mentioning that Attention-RPN and FSDetView requires GT bounding boxes for all objects to train, thus, we generate the pseudo GT bounding boxes using the FamNet+RR with the feature extracted from the dot annotations of training images instead of peak locations.
% and our first stage of \Approach~training (as described in Sec.~\ref{sec:two_stage}).

On FSCD-147, our method significantly outperforms others with a large margin for object detection. For counting, compared to a density-based approach like FamNet, \Approach~achieves worse results in RMSE metric but with much better results in other counting metrics MAE, NAE, and SRE. FamNet+MLP seems to overfit to the exemplar boxes so it performs the worst in detection.
% Fig.~\ref{fig:compare_abs_n_rel_error_fscd_famnet} shows the mean absolute errors of \Approach~and FamNet for different subsets of images, grouped by the number of objects contained in the images. For images with more than 200 objects, the average error of \Approach~is higher than that of FamNet, and this explains why FamNet is better in terms of RMSE. 
%This suggests that \Approach~produces more stable counting result across images in general but struggles if number of objects per images is  high ($>$200 objects).

% \begin{figure}[t]
%   \subfloat[]{
% 	\begin{minipage}[c]{
% 	   0.24\textwidth}
% 	   \centering
% 	   \label{fig:gaussian_height}
% 	   \includegraphics[width=1\textwidth]{images/gaussian_height.pdf}
% 	\end{minipage}}
%   \subfloat[]{
% 	\begin{minipage}[c]{
% 	   0.24\textwidth}
% 	   \centering
% 	   \label{fig:gaussian_width}
% 	   \includegraphics[width=\textwidth]{images/gaussian_width.pdf}
% 	\end{minipage}}
%   \subfloat[]{
% 	\begin{minipage}[c]{
% 	   0.24\textwidth}
% 	   \centering
% 	   \label{fig:laplace_height}
% 	   \includegraphics[width=1\textwidth]{images/laplace_height.pdf}
% 	\end{minipage}}
%   \subfloat[]{
% 	\begin{minipage}[c]{
% 	   0.24\textwidth}
% 	   \centering
% 	   \label{fig:laplace_width}
% 	   \includegraphics[width=\textwidth]{images/laplace_width.pdf}
% 	\end{minipage}}

% \caption{Overlap between train and val sets of predicted object height/width with Gaussian/Laplace distribution
% }
% \label{fig:overlap_distribution}
% \end{figure}

\begin{figure}[t]
% \centering
\includegraphics[width=1.0\textwidth]{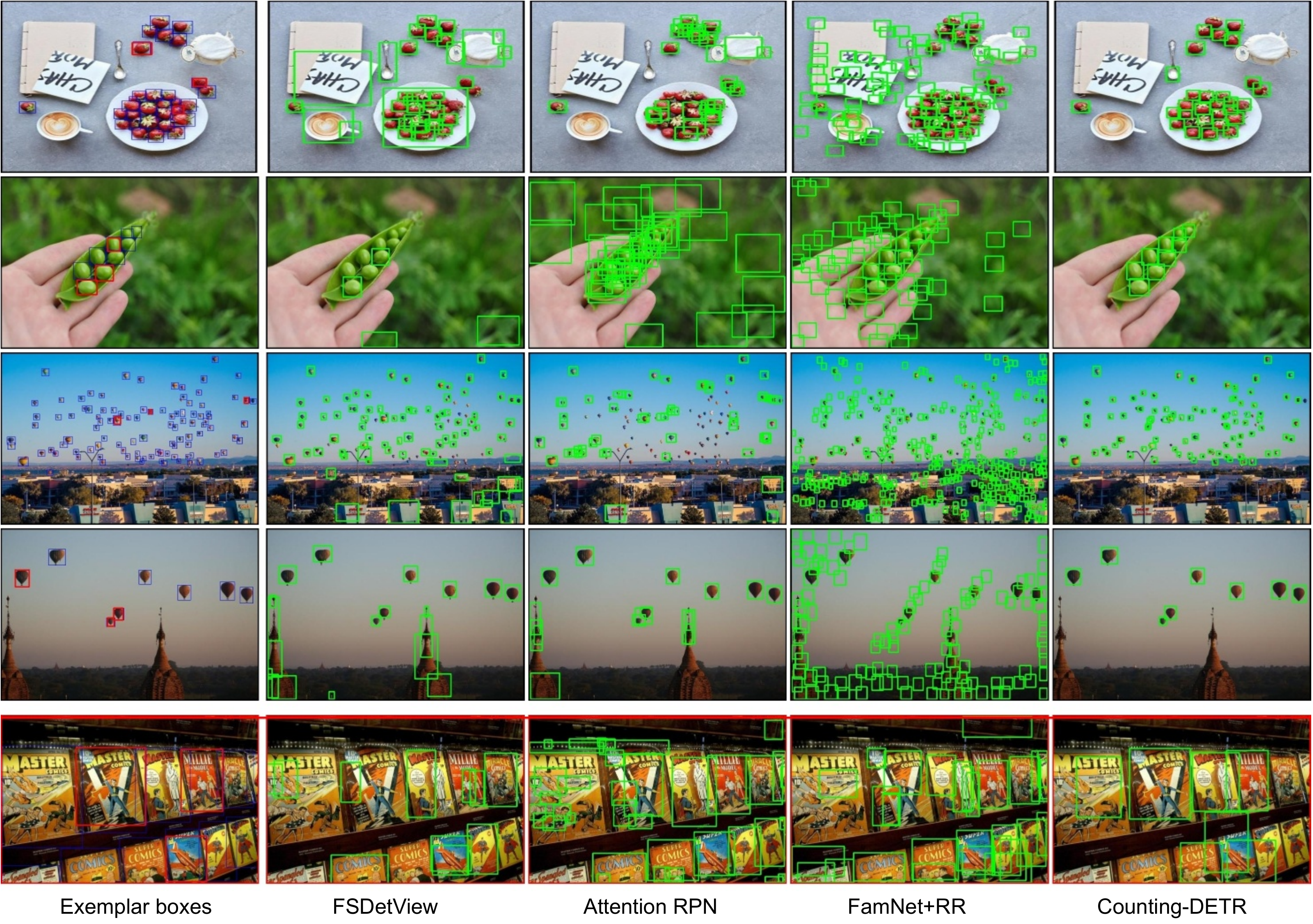}
\vskip -0.1in
\caption{{\bf Qualitative comparison}. Each row shows an example with the exemplars (red) and GT bounding boxes (blue) on the first column. The first four rows show the superior performance of ours over the others, while the last row is a failure case. In the first row, our method can distinguish between the target class and other foreground classes, while other methods confuse between different foreground classes. In the next three rows, exemplar objects either share similar color with environment or contain many background pixels. These conditions lead to either under detect or over detect where other methods are unsure about if detected objects are foreground or background. The last row shows a failure case for all methods due to the huge variation in the scale, distortion, and truncation of the objects.
% \mhoai{The name of the method has changed. It is also not clear whether it is working or not. It is not obvious what this figure shows and whether the method is working or not. Need to highlight the success better. }
}
\vspace{-10pt}
\label{fig:qualitative}
\end{figure}

%<<<<<<< HEAD
%In FSCD-LVIS, our method outperforms all others in both detection and counting. This is because the images in FSCD-LVIS are much more complicated than those in FSCD-147, having multiple object classes per image and significant variation in object sizes and shapes. Moreover, the class of interest is usually packed and occluded by other classes, so that density maps cannot be reliably predicted as shown in Fig.~\ref{fig:famnet_on_lvis}.
%=======
On FSCD-LVIS, our method outperforms all others for both detection and counting tasks. 
This is because the image in FSCD-LVIS is much more complicated than those in FSCD-147, i.e., multiple object classes per image and significant differences in object size and shape. Also, the class of interest is usually packed and occluded by other classes, so the density map cannot be reliably predicted as shown in Fig.~\ref{fig:famnet_on_lvis}. More interestingly, we also evaluate the performance of \Approach~and other baselines on a special test set of unseen classes of the FSCD-LVIS dataset to show their generalizability to unseen classes during training in Tab.~\ref{tab:LVIS_results_unseen}. It can be seen that our approach performs the best while the FamNet+RR performs the worst.
% Notably, our approach also achieves comparable results with the same version trained with the real GT bounding boxes provided by the dataset. This suggests that the quality of our pseudo bounding boxes approaches that of the GT bounding boxes.

% \Approach  also gain better results in detection which can use to interpret counting results. From Tab.~\ref{tab:LVIS_results}, density-base methods such as FamNet is not suitable for few shot object counting when the scenes are too complex and object's shape and size are too various which reduce ability to correlate between exemplar objects and image. 
% ....\khoi{Please write your observation and comments}

%\subsection{Qualitative Results}

Fig.~\ref{fig:qualitative} shows the qualitative comparison between our approach and the other methods, including FSDetView \cite{xiao2020few}, Attention-RPN \cite{fan2020few}, and FamNet \cite{m_Ranjan-etal-CVPR21}+RR. Our method can successfully detect the objects of interest while other methods cannot, as shown in the first four rows of Fig.~\ref{fig:qualitative}. The last row is a failure case for all methods, due to object truncation, perspective distortion, and scale variation.

\section{Conclusions}
\label{sec:conclusion}

We have introduced a new task of few-shot object counting and detection that shares the same supervision with few-shot object counting but additionally predict object bounding boxes. To address this task, we have collected two new datasets,
% The former extends from FSC-147 with bounding box annotation in test and val sets and the latter is adapted from the LVIS dataset with more object classes, higher variance of object size and shape, and more cluttered scene than those of FSCD-147. 
 adopted a two-stage training strategy to generate pseudo bounding boxes for training,  and developed a new uncertainty-aware few-shot object detector to adapt to the imperfection of pseudo label. Extensive experiments on the two datasets demonstrate that the proposed approach outperforms  strong baselines adapted from few-shot object counting and few-shot object detection. 
 
 %We have also carefully analyzed the cases where \Approach~succeeds and fails for promoting future work.

% that detect and counting at the same time which can be used to reduce time and effect to annotate objects in interested class. We created 2 new datasets which are FSCD-147 and FSCD-LVIS to benchmark methods for this new task. These two datasets are challenging due to complex scene, variation in shape, size of objects and number of objects per image. We proposed 2 stage-training approach to generate pseudo boxes from dot annotations to train our model.Also, we proposed new uncertainty-aware loss that alleviate the imperfection pseudo boxes. Extensive results on two datasets demonstrate proposed approach perform competitively against all baseline. Our future work might be integrate density-based with point-based approaches to further improve counting results in images with very large number of objects.

\clearpage
\bibliographystyle{splncs04}
\bibliography{egbib}
\end{document}